\documentclass{article}

\usepackage[final]{neurips_2023}

\usepackage[utf8]{inputenc}  
\usepackage[T1]{fontenc}     
\usepackage[table,xcdraw]{xcolor}  
\usepackage{hyperref}        
\usepackage{url}             
\usepackage{booktabs}        
\usepackage{amsfonts}        
\usepackage{nicefrac}        
\usepackage{microtype}       
\usepackage{xcolor}          

\usepackage{multirow}        

\usepackage{authblk}         
\usepackage{graphicx}        
\usepackage{svg}             
\usepackage{cleveref}        
\usepackage{scalefnt}        
\usepackage{chngcntr}        
\usepackage{float}           
\usepackage{pifont}          

\title{ShapefileGPT: A Multi-Agent Large Language Model Framework for Automated Shapefile Processing}

\author[1]{Qingming Lin}
\author[2]{Rui Hu}
\author[3]{Huaxia Li}
\author[2]{Sensen Wu}
\author[4]{Yadong Li}
\author[1]{Kai Fang}
\author[1]{Hailin Feng}
\author[2]{\\ Zhenhong Du}
\author[1,2]{Liuchang Xu$^{\dagger}$}

\affil[1]{School of Mathematics and Computer Science, Zhejiang Agriculture and Forestry University}

\affil[2]{School of Earth Sciences, Zhejiang University}

\affil[3]{Xiaohongshu}

\affil[4]{Baichuan Inc.}

\affil[$\dag$]{Corresponding author}

\begin{document}

\maketitle

\newcommand{\myfootnote}[1]{%
  \begingroup
  \renewcommand\thefootnote{}\footnotetext{#1}%
  \endgroup
}

\myfootnote{$^{1}$ E-mails: linqingming@stu.zafu.edu.cn}
\myfootnote{$^{\dagger}$ Corresponding author. E-mails: xuliuchang@zafu.edu.cn}

\begin{abstract}
Vector data is one of the two core data structures in geographic information science (GIS), essential for accurately storing and representing geospatial information.
Shapefile, the most widely used vector data format, has become the industry standard supported by all major geographic information systems.
However, processing this data typically requires specialized GIS knowledge and skills, creating a barrier for researchers from other fields and impeding interdisciplinary research in spatial data analysis.
Moreover, while large language models (LLMs) have made significant advancements in natural language processing and task automation, they still face challenges in handling the complex spatial and topological relationships inherent in GIS vector data.
To address these challenges, we propose ShapefileGPT, an innovative framework powered by LLMs, specifically designed to automate Shapefile tasks.
ShapefileGPT utilizes a multi-agent architecture, in which the planner agent is responsible for task decomposition and supervision, while the worker agent executes the tasks.
We developed a specialized function library for handling Shapefiles and provided comprehensive API documentation, enabling the worker agent to operate Shapefiles efficiently through function calling.
For evaluation, we developed a benchmark dataset based on authoritative textbooks, encompassing tasks in categories such as geometric operations and spatial queries.
ShapefileGPT achieved a task success rate of 95.24\%, outperforming the GPT series models.
In comparison to traditional LLMs, ShapefileGPT effectively handles complex vector data analysis tasks, demonstrating superior spatial data understanding and analytical capabilities, and overcoming the limitations of traditional models in spatial reasoning.
This breakthrough opens new pathways for advancing automation and intelligence in the GIS field, with significant potential in interdisciplinary data analysis and application contexts.
\end{abstract}
\section{Introduction}
The integration of geographic information science (GIS) with artificial intelligence has propelled the development of geospatial artificial intelligence (GeoAI) \cite{wang2024researchspatialdataintelligent}.
During this process, vector data plays a crucial role in responsibility of representing spatial information and describing geographic objects along with their spatial relationships \cite{temporary-citekey-5333}.
Shapefile, the most commonly used vector data format, and although it offers broad compatibility and flexibility, efficiently handling Shapefiles typically requires specialized GIS knowledge and skills.
As an interdisciplinary field, GIS has seen applications in urban planning, environmental science, agriculture, public health, and many other fields.
For researchers or professionals in these fields, limited GIS expertise often becomes a significant barrier to using Shapefiles for spatial data analysis.
While the Shapefile format offers great compatibility and flexibility, its manipulation and analysis generally rely on professional GIS software like ArcGIS or QGIS, which imposes a steep learning curve for non-GIS users.
Lowering the technical barriers for conducting spatial analysis of vector data has thus become a key challenge in advancing the widespread adoption and development of GeoAI.

In recent years, large language models (LLMs) have made significant strides in processing both text and structured data, demonstrating impressive capabilities in automated data processing \cite{openai2024gpt4technicalreport, 10.1145/3616855.3635752, zha2023tablegptunifyingtablesnature}.
However, LLMs face considerable challenges when addressing complex spatial tasks in GIS \cite{xu2024evaluatinglargelanguagemodels, 10.1145/3615886.3627745}.
Specifically, while current GPT models can generate code and automate routine tasks, the accuracy and reliability of the generated code are not always assured \cite{NEURIPS2023_43e9d647}.
These limitations become even more pronounced when dealing with the intricate spatial and topological relationships unique to the GIS, where the generated code often fails to meet the requirements of professional applications.
This suggests that developing an LLM framework specifically tailored for Shapefiles would not only better address the complex demands of GIS, improving the accuracy and automation of task handling, but also offer researchers from interdisciplinary fields an accessible tool for effectively managing spatial data.

Meanwhile, tool-calling technology is particularly crucial in the GIS domain.
As LLMs continue to evolve, they have shown remarkable progress in reasoning and knowledge integration.
For example, through techniques like chain-of-thought~\cite{NEURIPS2022_9d560961} and reinforcement learning~\cite{bai2022traininghelpfulharmlessassistant}, advanced models such as OpenAI's o1 model excel in logical reasoning, applying domain-specific knowledge, and natural language processing \cite{zhong2024evaluationopenaio1opportunities}.
Nevertheless, without deep integration into external specialized systems, the potential of these models in the GIS domain remains largely untapped.
Enabling LLMs to interact with external systems through tool-calling can significantly expand their utility in specialized contexts, especially in handling complex vector data tasks \cite{qin2023toolllmfacilitatinglargelanguage}.
Currently, efficient solutions integrating LLMs with Shapefiles in GIS are lacking, presenting a novel entry point for our research.

\begin{figure}
  \centering
  \includegraphics[width=\textwidth]{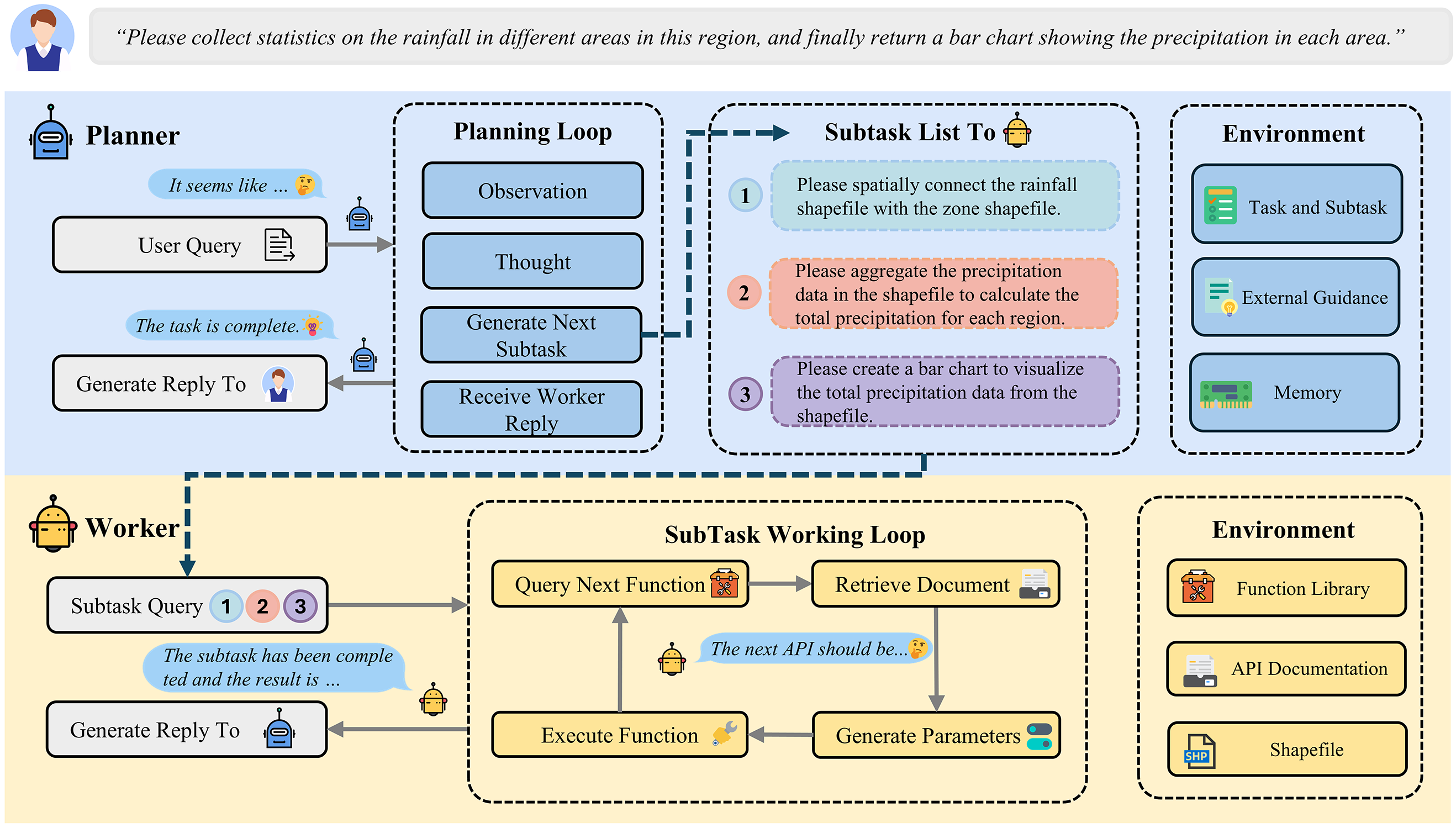}
  \caption{ShapefileGPT consists of a planner agent and a worker agent. The planner agent interprets user queries and decomposes them into subtasks, while the worker agent executes these subtasks by selecting appropriate functions from a predefined function library to perform Shapefile-related operations.}
  \label{fig:overview}
\end{figure}

To address these challenges, we propose ShapefileGPT, a GIS agent system specifically designed for the Shapefile format.
ShapefileGPT simplifies user interaction by enabling direct handling of Shapefile tasks through natural language, thereby lowering the technical barrier and significantly improving efficiency.
As illustrated in Fig.~\ref{fig:overview}, ShapefileGPT employs a multi-agent architecture consisting of a planner agent and a worker agent. The planner agent interprets user commands and decomposes them into subtasks, while the worker agent uses a specialized function library to execute these subtasks efficiently.
This architecture enables ShapefileGPT to automate complex vector data tasks.
Additionally, we developed a Shapefile task dataset to evaluate the performance of ShapefileGPT against other LLMs in Shapefile data processing. In our experiments, ShapefileGPT achieved a task success rate of 97.62\%, far surpassing GPT-4o's 45.24\%.
These results highlight ShapefileGPT's accuracy in vector data spatial analysis, effectively addressing the spatial understanding limitations of traditional LLMs.

The introduction of ShapefileGPT not only addresses the existing challenges in GIS operations but also establishes a new paradigm for automating spatial data analysis in GIS, advancing the development of GeoAI.
By lowering the barriers to GIS technology, researchers from interdisciplinary fields can readily leverage Shapefile data for complex spatial analyses, fostering cross-disciplinary collaboration and innovation.
Our main contributions are as follows:
\begin{itemize}
    \item We introduce ShapefileGPT, an agent framework powered by LLMs to handle Shapefile tasks, significantly enhancing the automation capabilities of LLMs in vector data processing. This reduces the technical barriers for users, especially benefiting non-GIS professionals.
    \item We developed a Shapefile dataset and systematically evaluated the performance of ShapefileGPT against current GPT models in vector data processing tasks.
    \item We designed a specialized Shapefile processing function library that integrates tool-calling with LLMs, addressing the inherent limitations of LLMs in vector data processing and providing an efficient, accurate processing mechanism.
\end{itemize}

\section{Related Works}

\textbf{Large Language Model Agents}
The landscape of large language model (LLM) agents has rapidly evolved, achieving a series of breakthroughs.
An LLM agent is defined as a framework composed of three components: a brain, perception, and action \cite{xi2023risepotentiallargelanguage}. Agent architectures can be classified as single-agent or multi-agent, depending on the number of agents involved \cite{masterman2024landscapeemergingaiagent}.
In the ReAct framework \cite{yao2023reactsynergizingreasoningacting}, the agent first reasons about the task and then executes actions based on that reasoning, demonstrating greater effectiveness than traditional methods, such as zero-shot prompting, where the model acts without prior task-specific training.
Building on ReAct, RAISE~\cite{liu2024llmconversationalagentmemory} introduces a memory mechanism to enhance the agents' ability to retain and utilize contextual information.
Reflexion~\cite{shinn2023reflexionlanguageagentsverbal} employs self-reflection in a single-agent system, leveraging an LLM evaluator and metrics like success state and memory to provide targeted feedback, significantly improving overall task performance.
Moreover, ~\cite{guo2024embodiedllmagentslearn} underscores the crucial role of a lead agent in enhancing the effectiveness of embodied LLM agent teams.
AgentVerse~\cite{chen2023agentversefacilitatingmultiagentcollaboration} demonstrates how structured phases in group planning enhance agents' reasoning and problem-solving abilities.

\textbf{GIS Agent Framework}
In the GIS field, an increasing number of studies are focusing on how to utilize LLM agents to automate geospatial data processing, reducing human intervention and improving analysis efficiency \cite{doi:10.1080/17538947.2024.2353122}.
Several LLM-based automation frameworks have demonstrated significant progress in GIS applications.
LLM-Geo~\cite{doi:10.1080/17538947.2023.2278895} has shown great potential in the study of autonomous GIS.
The system performs automatic data generation and autonomous task execution through LLMs, allowing for automatic map generation, spatial data aggregation, and result visualization.
MapGPT~\cite{doi:10.1080/15230406.2024.2404868} is an intelligent mapping framework.
It uses natural language processing to understand user requirements and calls various mapping tools to generate map elements, significantly simplifying the map-making process, while offering users greater creative control.
Additionally, research has extracted knowledge from Google Earth Engine (GEE) workflow scripts related to geographic analysis models \cite{doi:10.1080/17538947.2024.2398063}.
The framework extracts descriptive and procedural knowledge from complex GEE scripts related to geographic analysis and packages this knowledge into reusable templates, enabling sharing and application in various geographic modeling environments.
POI GPT~\cite{isprs-archives-XLVIII-4-W10-2024-113-2024} uses named entity recognition (NER) combined with LLMs to extract precise POI locations from text, reducing the cost and time traditionally required to extract spatial information from text data like social media posts.
A study~\cite{Xu2024GenAIpoweredMP} explored the use of multi-agent systems in intelligent transportation, utilizing retrieval-augmented generation (RAG) technology to improve the efficiency of smart city mobility applications.
GPT4GEO~\cite{Roberts2023GPT4GEOHA} explored the capabilities of GPT-4 in geospatial tasks such as route planning, disaster management, and supply-chain analysis.

\textbf{Tool-augmented Large Language Models}
Tool-augmented Large Language Models enable LLMs to connect with external tools, effectively overcoming inherent limitations.
By integrating resources such as search engines for external knowledge access and calculators to enhance mathematical capabilities, LLMs can also be utilized for repetitive daily tasks \cite{komeili-etal-2022-internet, heyueya2023solvingmathwordproblems, NEURIPS2023_d842425e}.
One significant challenge with LLMs is their black-box nature, along with the issue of hallucinations, where models generate inaccurate information.
To address this, research has focused on enhancing retrieval capabilities, enabling LLMs to generate citation-backed content, thus increasing the trustworthiness of their output \cite{sun2024verifiabletextgenerationevolving}.
Compared to traditional LLMs, LLM agents demonstrate enhanced intelligence, particularly when integrated with external tools.
These tool-augmented agents can handle more complex tasks.
For example, the open-source project AutoGPT operates as a fully autonomous system, executing tasks without user intervention \cite{bonatti2024windowsagentarenaevaluating}.
Additionally, Navi, a multi-modal agent capable of processing various input types, simulates human operations within the Windows operating system \cite{githubGitHubSignificantGravitasAutoGPT}.

Inspired by the above research, we propose ShapefileGPT.
While existing studies have demonstrated the vast potential of LLMs in GIS applications, they primarily focus on high-level data generation or the automation of specific tasks, often lacking support for concrete vector data operations.
Additionally, current GIS automation frameworks are primarily designed for professional GIS users, leaving researchers from non-GIS fields to face high technical barriers when using these tools.
ShapefileGPT addresses these gaps by enhancing LLMs' ability to understand and analyze vector data, making it applicable not only to specific use cases, but also to a broad range of spatial data analysis tasks.
Moreover, ShapefileGPT lowers the technical barriers for non-specialists by enabling natural language interaction, thereby facilitating more widespread use of GIS data processing in interdisciplinary collaborations.
\section{Methodology}
\subsection{Overview}
\label{sec:methodology:overview}
ShapefileGPT enables the execution of spatial analysis tasks on Shapefiles using natural language.
It takes both the full Shapefile and the user’s task as input, progressively analyzes and processing the vector data, and ultimately saves the results as images, tables, or Shapefile files.
The system employs a division of labor between the planner agent and the worker agent.
The planner agent serves as the decision-making hub, receiving user instructions and breaking them down into subtasks, while the worker agent executes these specific subtasks.
The agents communicate through internal APIs and collaborate to automate task processing.

The planner agent is equipped with advanced observation, reasoning, and memory capabilities, allowing it to accurately interpret user requirements and intelligently decompose them into detailed subtasks.
During task execution, the planner agent continuously monitors progress, formulates and adjusts subtasks based on real-time conditions, and guides the worker agent in its execution.
As the worker agent completes tasks and reports the results, the planner agent updates its memory to optimize future task management, iterating until the entire task is completed and the final result is returned to the user.

The worker agent handles specific data processing tasks by invoking the specialized Shapefile function library through APIs, ensuring precision and efficiency.
It conducts detailed analyses of each subtask, executes them sequentially, and provides real-time feedback to the planner agent.

\subsection{Enhancing LLMs with Function Calling}
\label{sec:methodology:function-calling}
We enable LLMs to execute actual Shapefile tasks through function calling.
Function Calling is a mechanism that enhances interaction between LLMs and external programs, enabling the model to invoke predefined functions while generating text.
This mechanism enables the execution of complex tasks by allowing LLMs to not only generate natural language but also interact with external programs, databases, and APIs.
It supports complex computations, real-time data access, and specialized task execution.

In contrast to function calling, online LLM models typically rely on code generation to execute specific tasks.
For example, in OpenAI's online GPT-4o service, the model first parses the task requirements and generates the corresponding code, then executes the code in a virtual environment, validates its effectiveness based on the results, and finally returns the outcome to the user.
However, our experiments demonstrate that GPT series models (such as GPT-4 and GPT-4o) often fail to accurately complete Shapefile tasks.
To address these issues, we designed ShapefileGPT using a function-calling mechanism, where functions are specifically crafted, and LLMs are guided to invoke them correctly, enabling more accurate execution of Shapefile tasks.

To enhance the accuracy of Shapefile task processing, we designed a set of functions specifically tailored for Shapefile handling in ShapefileGPT, developing a function library with accompanying APIs and detailed documentation.
The API documentation outlines the usage and parameter rules for each function, helping ShapefileGPT select the appropriate API for task execution.
We developed 27 functions centered around the Shapefile data structure, covering both basic tabular data operations (e.g., renaming, filtering, and adding fields) and vector data operations (e.g., spatial joins, buffer generation, clipping, and geometry transformations).
These functions equip the LLM with essential geometric and spatial analysis capabilities, enabling it to effectively handle Shapefile data.

During task execution, ShapefileGPT gradually analyzes the task based on user instructions and real-time feedback, selecting the appropriate API from the function library.
For example, when handling spatial overlay tasks, ShapefileGPT first assesses the necessary geometric processing steps, then calls the relevant API as the task progresses, avoiding redundant operations.
This API selection process mirrors the workflow of GIS professionals, ensuring task accuracy by methodically processing spatial data step-by-step.
To further enhance efficiency, we categorized the APIs by the stages of Shapefile task processing into three main types: "Data Reading", "Processing and Analyzing Data", and "Saving Data", with multiple subcategories under each.
This categorization narrows the search scope for ShapefileGPT during API selection, improving both the efficiency and accuracy of function calling.

\begin{figure}[H]
  \centering
  \includegraphics[width=\textwidth]{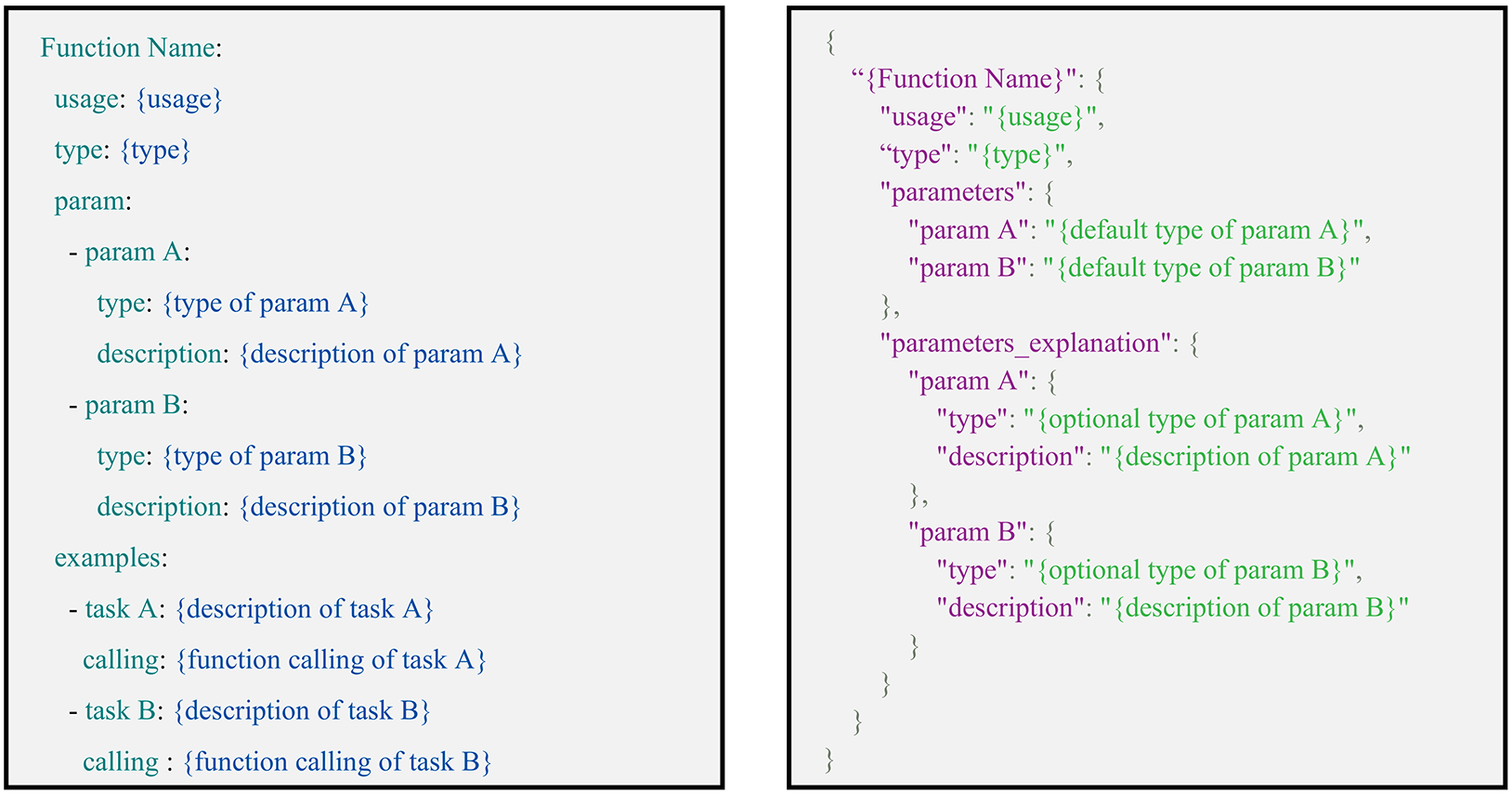}
  \caption{The design of the ShapefileGPT function library documentation. The YAML format (left) is concise and used for LLM context guidance, while the JSON format (right) is structured for external system interaction and result validation.}
  \label{fig:methodology:function_format}
\end{figure}

Our API documentation provides throughout details on each API's function name, parameter definitions and types, usage, and examples.
As shown in Fig.~\ref{fig:methodology:function_format}, the documentation is stored in both YAML and JSON formats: YAML for its simplicity and readability, which allows it to be embedded in the LLM’s context for real-time guidance, and JSON for verifying and evaluating the accuracy of the LLM’s function calling.
Although JSON has a more complex syntax, which increases token usage when embedded in context, its clear key-value structure makes it ideal for interacting with external systems and verifying results.

\subsection{Multi-Agent Framework}
\label{sec:method:multi-agent}
\noindent\textbf{Multi-Agent architecture}
Our proposed ShapefileGPT adopts a vertical multi-agent architecture.
While a single-agent architecture performs well for simple, straightforward tasks, it exhibits limitations when confronted with complex reasoning, such as an inability to process tasks in parallel and a tendency to generate hallucinations \cite{masterman2024landscapeemergingaiagent}.
In the vertical multi-agent architecture, one agent acts as the leader, responsible for overall planning and monitoring sub-tasks, while the other agents serve as executors to carry out specific tasks.
This architecture is particularly well-suited for complex multi-step tasks, especially in vector data analysis, as it simplifies the execution of each step while maintaining logical clarity.

\begin{figure}[H]
  \centering
  \includegraphics[width=\textwidth]{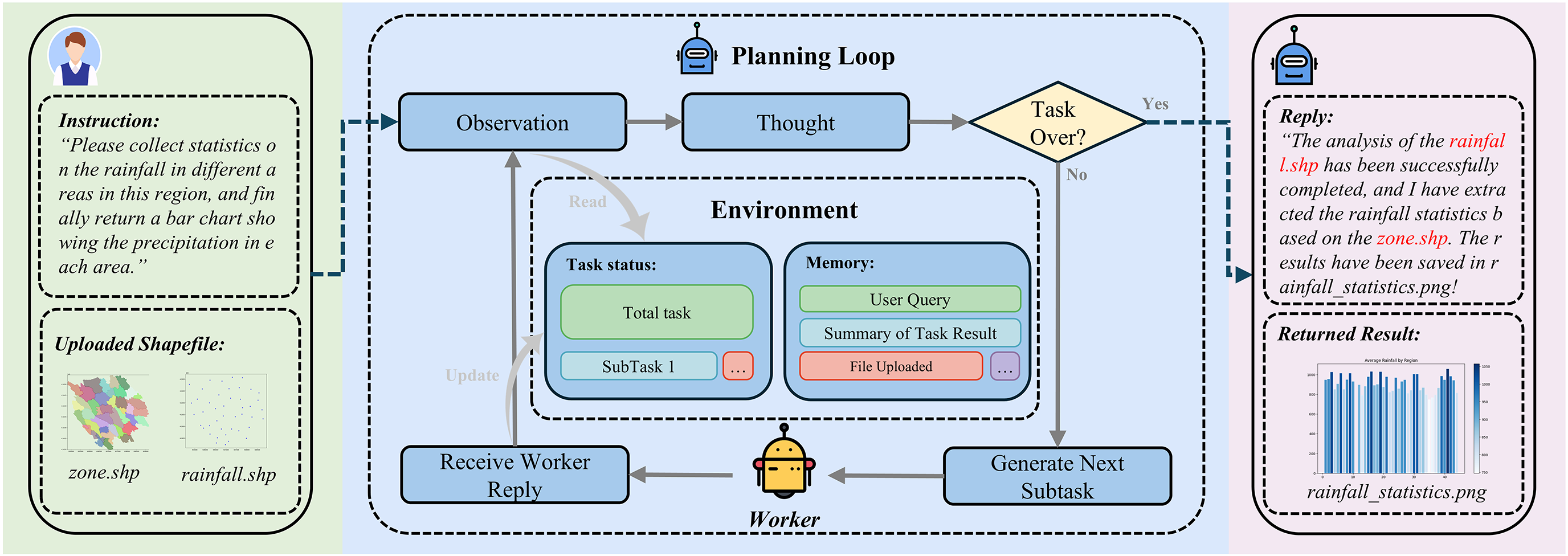}
  \caption{The planning loop in ShapefileGPT's multi-agent framework. The planner agent continuously monitors task progress, formulates new subtasks based on real-time conditions, and updates the task environment to ensure seamless task execution.
}
  \label{fig:methodology:planner}
\end{figure}

\noindent\textbf{Planner Agent}
In our architecture, the planner functions as the leader, and the worker operates as the executor.
The planner’s role is to break down the user's instructions into multiple subtasks and assign them to the worker for execution.
As shown in Fig.~\ref{fig:methodology:planner}, after the user uploads the Shapefile and task instructions, the system initializes the planner's work environment, which records task progress, current task status, and the planner’s memory state, including information about previously executed tasks.
The planner conducts task planning through a planning loop, where each cycle represents the lifecycle of a subtask, from breakdown to completion.
In each loop, the planner first observes the current task state and determines whether the task has been completed.
If the task is not complete, the planner generates a new sub-task and assigns it to the worker for execution.
Once the planner receives results from the worker, it updates the task environment to ensure that subsequent tasks are planned and executed correctly.

\begin{figure}[H]
  \centering
  \includegraphics[width=\textwidth]{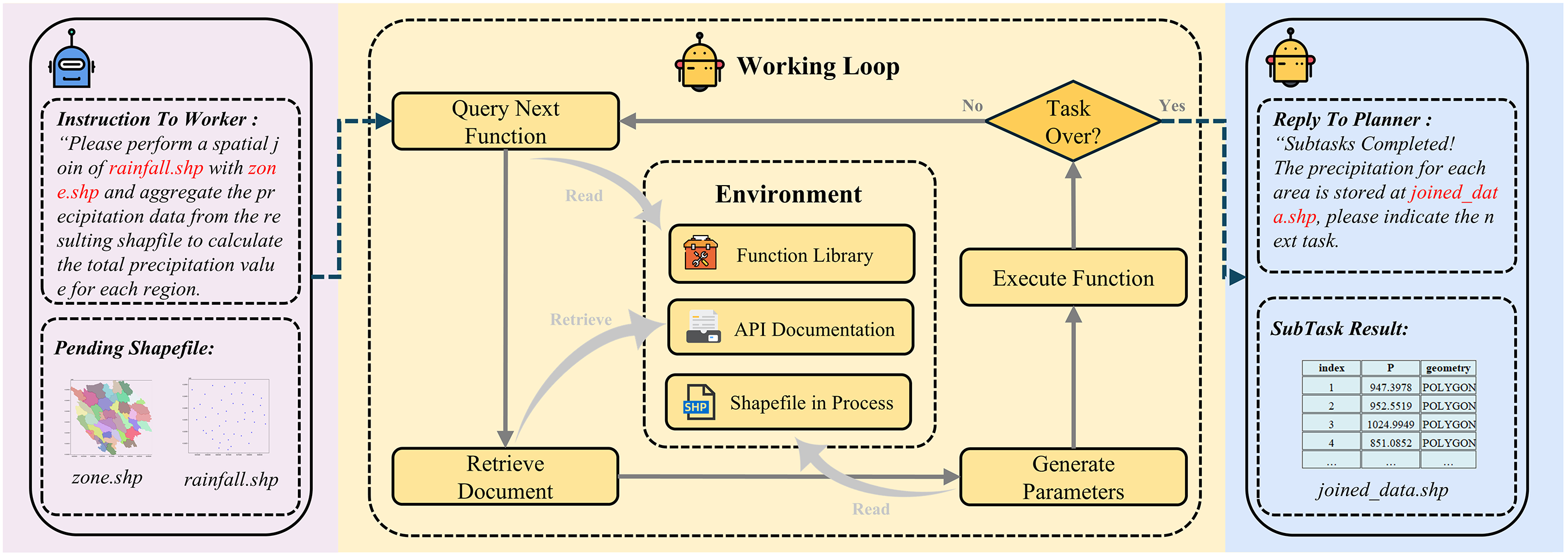}
  \caption{The worker loop in ShapefileGPT's multi-agent framework. The worker agent selects the appropriate API from a specialized function library to execute subtasks, ensuring precise handling of Shapefile data.
}
  \label{fig:methodology:worker}
\end{figure}

\noindent\textbf{Worker Agent}
The worker, operating as the executor, follows the workflow outlined in the working loop shown in Fig.~\ref{fig:methodology:worker}.
Upon receiving a task from the planner, the worker initiates execution within its environment.
This environment consists of the function library and the provided API documentation.
The function library provides the names and functional descriptions of each API, guiding the worker in selecting the most suitable one.
The API documentation details the parameter rules for each API, ensuring the worker correctly configures the necessary parameters for API calls.
Additionally, the worker's environment contains descriptive information about the Shapefile to be processed, which is essential for vector data tasks.
During execution, the worker must understand the geometry type, field names, and attribute table information of the current Shapefile to prevent errors, such as referencing non-existent columns or fields during function calls.
This ensures both task accuracy and reliability.

Each cycle of the working loop encompasses the entire process, from selecting an API to executing it.
After receiving a task, the worker selects the appropriate API based on the documentation and generates a function call with the correct parameters.
ShapefileGPT’s backend executes these functions within a secure, isolated sandbox environment, ensuring that task execution neither interferes with other processes nor compromises data security.
Upon task completion, the worker evaluates the result to determine whether the task is complete and reports the outcome to the planner.
This feedback allows the planner to dynamically update the task environment, ensuring the smooth progression of subsequent tasks.

\subsection{Task Datasets}
\label{sec:method:datasets}
\begin{figure}[H]
  \centering
  \includegraphics[width=0.7\linewidth]{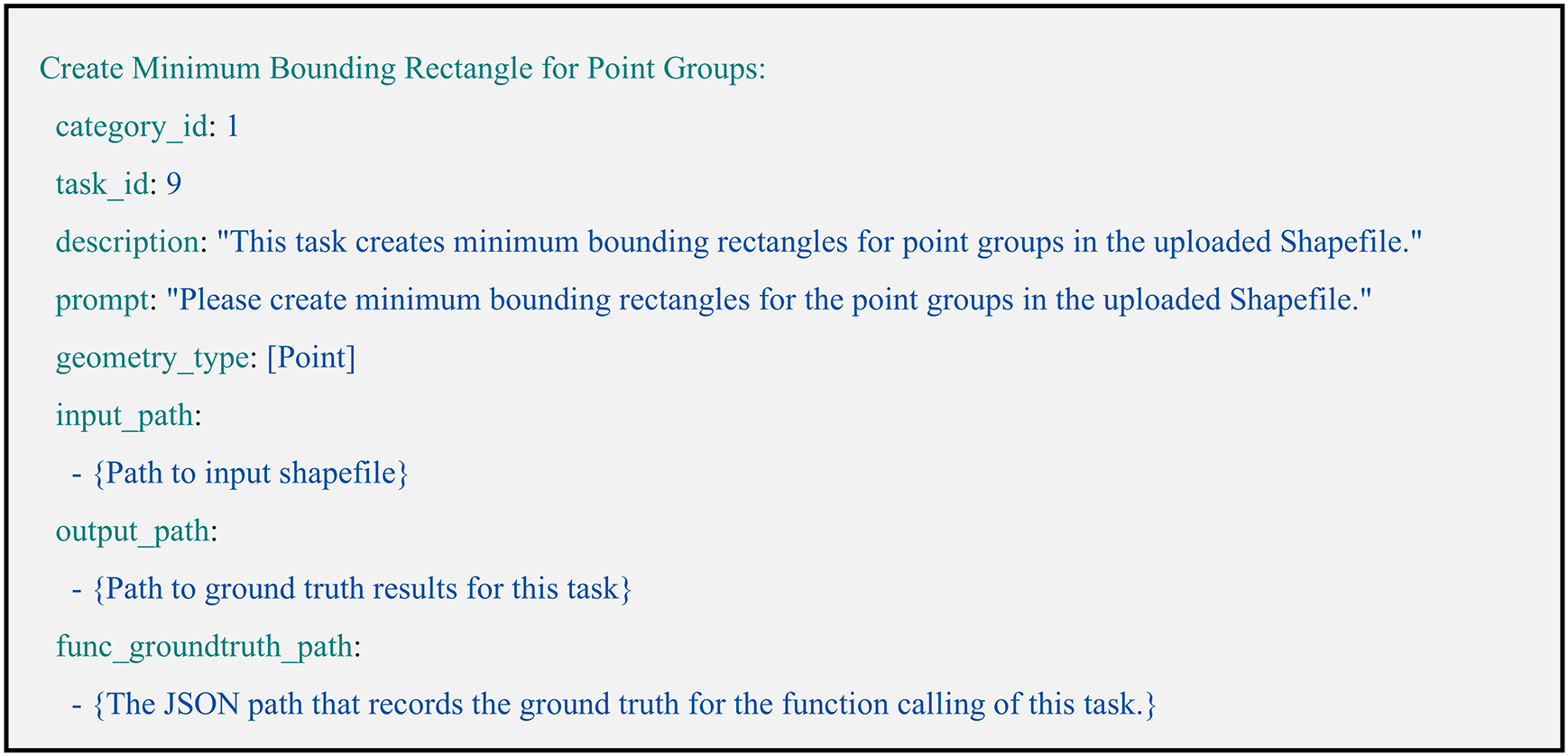}
  \caption{Example of a task from the Shapefile task dataset. This example demonstrates a geometric operation task, which involves creating minimum bounding rectangles for point groups in a Shapefile.}
  \label{fig:methodology:function_example}
\end{figure}

To evaluate the performance of online large language models against our designed ShapefileGPT, we constructed a Shapefile task dataset to test both models under identical conditions.
Building on the spatial analysis case from \cite{temporary-citekey-5333, 2017-ak}, we developed a standardized dataset for Shapefile tasks.
Each task includes a structured definition, consisting of a task ID, geometry type, category, description, input and output file paths, and user prompts.
An example of a task is shown in Fig.~\ref{fig:methodology:function_example}.

The proposed Shapefile task dataset consists of 42 tasks, with the distribution of tasks across categories shown in Table~\ref{tab:methodology:datasets}.
In GIS, spatial analysis tasks are typically divided into several categories.
Based on the spatial analysis classification outlined in \cite{temporary-citekey-5333}, we organized the tasks into the following main categories: geometric operation tasks, spatial query and calculation tasks, distance and direction tasks, grid operation tasks, and others.
Among these, geometric operations, spatial queries, and distance and direction calculations are closely related to the shape, position, and spatial relationships of spatial data, making them fundamental to spatial analysis.
As a result, we prioritized these three categories in the dataset.


\begin{table}[h]
\centering
\caption{Task categories in the Shapefile task dataset. }
\label{tab:methodology:datasets}

\begin{tabular}{lc}
\toprule
\textbf{Task category} & \textbf{Num}\\
\midrule %
Geometric Operations & 22 \\
Queries and Computations Operations & 7 \\
Distance and Direction Operations & 7 \\
Other Operations & 6 \\
Total & 42 \\
\bottomrule 
\end{tabular}

\end{table}

\label{sec:methodology:datasets}
In Shapefile tasks, vector data often requires geometric transformations, such as splitting or extending.
This step is particularly critical when dealing with complex geographic objects, as geometric transformations are essential for accurate data processing.
As result, geometric operations hold specific significance among the fundamental tasks.
To evaluate the performance of LLMs in executing Shapefile geometric operations and topological transformations, we developed 22 geometric operation tasks, as detailed in Table~\ref{tab:methodology:geometric-operations}.

\begin{table}[H]
\caption{Geometric operations included in the Shapefile task dataset. Each task involves manipulating or transforming the geometry of spatial features in a Shapefile.}
\label{tab:methodology:geometric-operations}

\centering
\small 
\resizebox{\linewidth}{!}{
\begin{tabular}{lp{0.7\textwidth}}
\toprule
\textbf{Task} & \textbf{Description} \\
\midrule
Calculate Geometry Length & Calculates the length of each geometric feature in the Shapefile, including the unique identifier (ID) and length, saved in a new Shapefile for further use. \\
Clip Layer & Clips the target Shapefile using a specified boundary. The result retains all features within the boundary and their attributes, saved as a new Shapefile. \\
Convert Geometry to Line Features & Converts geometric features in the Shapefile into line features, retaining original attributes, saved in a new Shapefile. \\
Convert Vertices of Lines or Polygons to Points & Converts vertices of line or polygon features into individual point features, saved as a new Shapefile. \\
Convert Line Features to Polygon Features & Converts line features into polygons, creating closed polygons from lines and retaining attributes, saved in a new Shapefile. \\
Overlay Analysis & Performs overlay analysis on two Shapefiles, identifying overlapping areas and retaining geometric and attribute information, saved as a new Shapefile. \\
Create Buffer Zones & Creates buffer zones of a specified distance around each feature, retaining related attributes, saved in a new Shapefile for further analysis. \\
Create Thiessen Polygons (Voronoi Diagram) & Creates Thiessen polygons (Voronoi diagrams) based on point features, dividing the spatial area, saved in a new Shapefile. \\
Create Minimum Bounding Rectangle for Point Groups & Generates minimum bounding rectangles for point groups, saving the smallest enclosing rectangle for each group in a new Shapefile. \\
Create Internal Buffer for Closed Lines & Creates internal buffers for line elements by generating buffer polygons within the line boundaries, ensuring buffer remains inside the original feature. \\
Analyze Disaster Impact Buffers & Generates multiple ring buffers to represent the impact zones of disasters on roads and areas, identifying affected roads and symbolizing risk levels. \\
Convert Vertices of Lines or Polygons to Points & Extracts vertices from line or polygon features and converts them into individual point features, saved with their unique identifiers in a new Shapefile. \\
Spatial Feature Analysis of Clustered Points & Analyzes the spatial distribution of clustered points by calculating centrality, distribution direction, and standard deviation to understand clustering patterns. \\
Add Coordinate Fields to Point & Adds X and Y coordinate fields to each point feature, saving the updated dataset with the original attributes in a new Shapefile. \\
Convert Start and End Coordinates of Point to Lines & Creates lines from the start and end points of each point feature, defining the lines by two points, saved in a new Shapefile. \\
Create Nearest Perpendicular Lines Between Line Features & Calculates and creates the nearest perpendicular line between line features, identifying start and end points of the perpendicular line. \\
Extract Overlapping Areas Between Polygons & Identifies and extracts overlapping areas between polygons, retaining attribute fields, saved in a new Shapefile for analysis. \\
Overlap Analysis of Polygon Features in Same Layer & Analyzes overlapping areas between polygon features, determining spatial relationships and attributes, saving the result in a new Shapefile. \\
Spatial Allocation of Points by Distance (Thiessen Polygons and Buffers) & Allocates spatial regions using Thiessen polygons or buffer zones based on point features, dividing areas by proximity, saved in a new Shapefile. \\
Create Thiessen Polygons for Polygon Features & Divides space based on proximity to polygon features, ensuring each area is closest to its corresponding feature, saved in a new Shapefile. \\
Split Polygons by Lines & Splits polygons using lines and saves the resulting new polygons in a Shapefile. \\
Create Mixed Thiessen Polygons for Points, Lines, and Areas & Generates Thiessen polygons for mixed point, line, and polygon features, indicating the nearest region to each feature, saved in a new Shapefile. \\
\bottomrule
\end{tabular}
}
\end{table}

Additionally, tasks such as coordinate system transformations, adding fields, and renaming fields are challenging to categorize.
Because these tasks are less related to the geometric properties of vector data, we grouped them under the category of other operations.

While current large language models excel in general knowledge and generalization, their performance in specialized tasks often falls short of expert-level proficiency.
Shapefiles contain vector data and associated attribute tables, which collectively define spatial features and their attributes.
Research suggests that LLMs demonstrate strong comprehension and manipulation abilities with structured tabular data \cite{10.1145/3616855.3635752, zha2023tablegptunifyingtablesnature}.
However, our experimental results with GPT indicate that LLMs do not handle vector data as effectively as general tabular data, particularly in tasks involving geometric transformations and spatial operations.
Therefore, enhancing LLMs' understanding of spatial geometric properties in vector data, including topological relationships, spatial adjacency, and directionality, is crucial for maximizing their reasoning capabilities in the GIS domain.
If an LLM can successfully operate on Shapefiles from the proposed dataset and produce the expected results, it would demonstrate a certain level of spatial analysis capability.

\begin{figure}
  \centering
  \includegraphics[width=\linewidth]{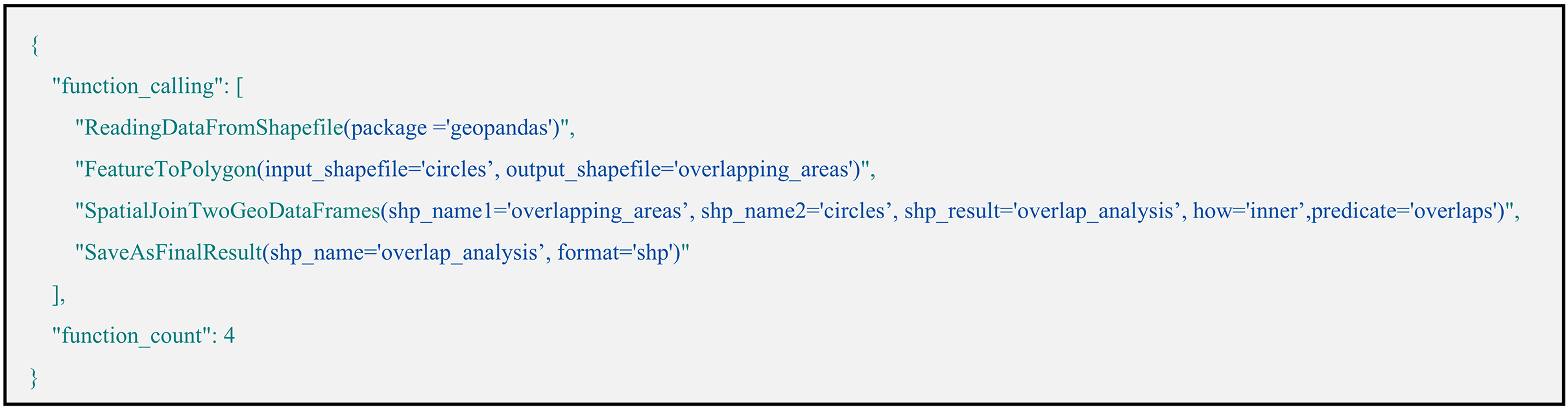}
  \caption{Example of function calling for Shapefile tasks. This example shows the process of function calling for Shapefile tasks within ShapefileGPT. It includes reading Shapefile data, converting features to polygons, performing a spatial join on overlapping areas, and saving the final result.}
  \label{fig:methodology:function_calling_example}
\end{figure}

Because ShapefileGPT executes Shapefile tasks through function calling, each task is assigned a ground truth path that records the optimal function calling process and parameter settings during execution.
An example of a spatial analysis function call is shown in Fig.~\ref{fig:methodology:function_calling_example}.
This establishes a clear standard and serves as the foundation for systematically evaluating ShapefileGPT’s function calling capabilities.

\section{Experiments}
\begin{figure}[h]
  \centering
  \includegraphics[width=0.7\linewidth]{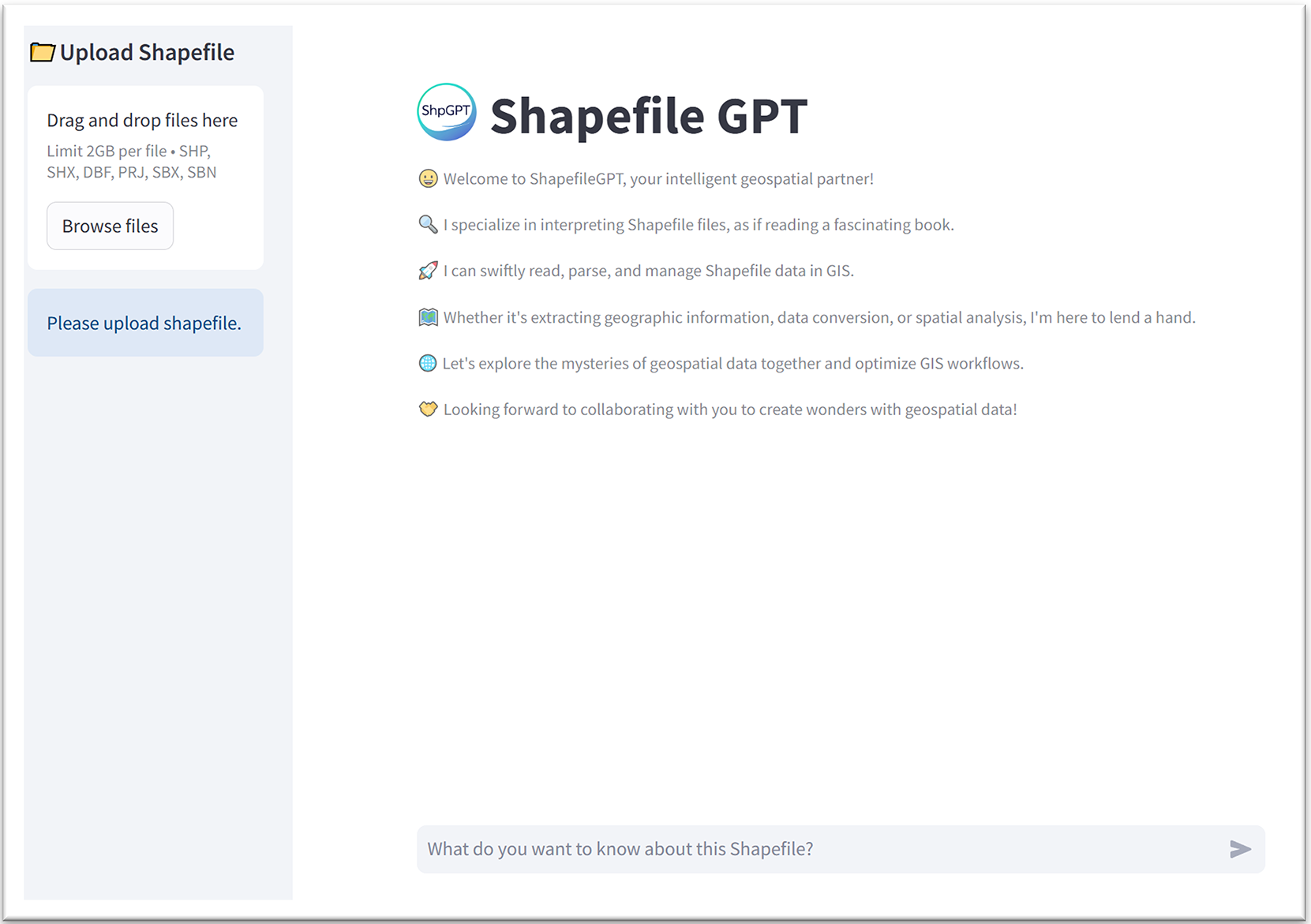}
  \caption{Interactive web interface of ShapefileGPT, developed using the Streamlit framework. The interface allows users to upload Shapefiles, submit natural language instructions, and receive processed outputs}
  \label{fig:experiment:web_interface}
\end{figure}

In this section, we present the experiments designed for our proposed ShapefileGPT and the corresponding results.
Specifically, Sec.~\ref{sec:experiment:setup} provides a detailed description of the experimental setup and the baseline models used for comparison.
Sec.~\ref{sec:experiment:comparison_with_gpt} presents the comparative results of ShapefileGPT and the GPT series models in executing Shapefile tasks.
Sec.~\ref{sec:experiment:comparison_with_different_config} describes the comparison experiments of ShapefileGPT with different configurations.
Sec.~\ref{sec:experiment:module_validation} validates the role of the Agent module through case studies.
Finally, Sec.~\ref{sec:experiment:ablation_planner} and ~\ref{sec:experiment:ablation_fewshot} present the results of ablation studies.
In addition to the experiments, we also developed an interactive web interface for ShapefileGPT using Python's Streamlit framework, enabling users to directly upload data and obtain results through the interface shown in Fig.~\ref{fig:experiment:web_interface}, further improving the system's ease of use and practicality.

\subsection{Setup}
\label{sec:experiment:setup}
\noindent\textbf{Datasets}
Our experiments are conducted using the task dataset introduced in Sec.~\ref{sec:methodology:datasets}.
This dataset encompasses a diverse range of Shapefile tasks, including geometric operations, spatial queries, and calculation tasks.
It is specifically designed to assess the capabilities of different models in handling complex vector data and performing related operations.

\begin{figure}
  \centering
  \includegraphics[width=0.7\linewidth]{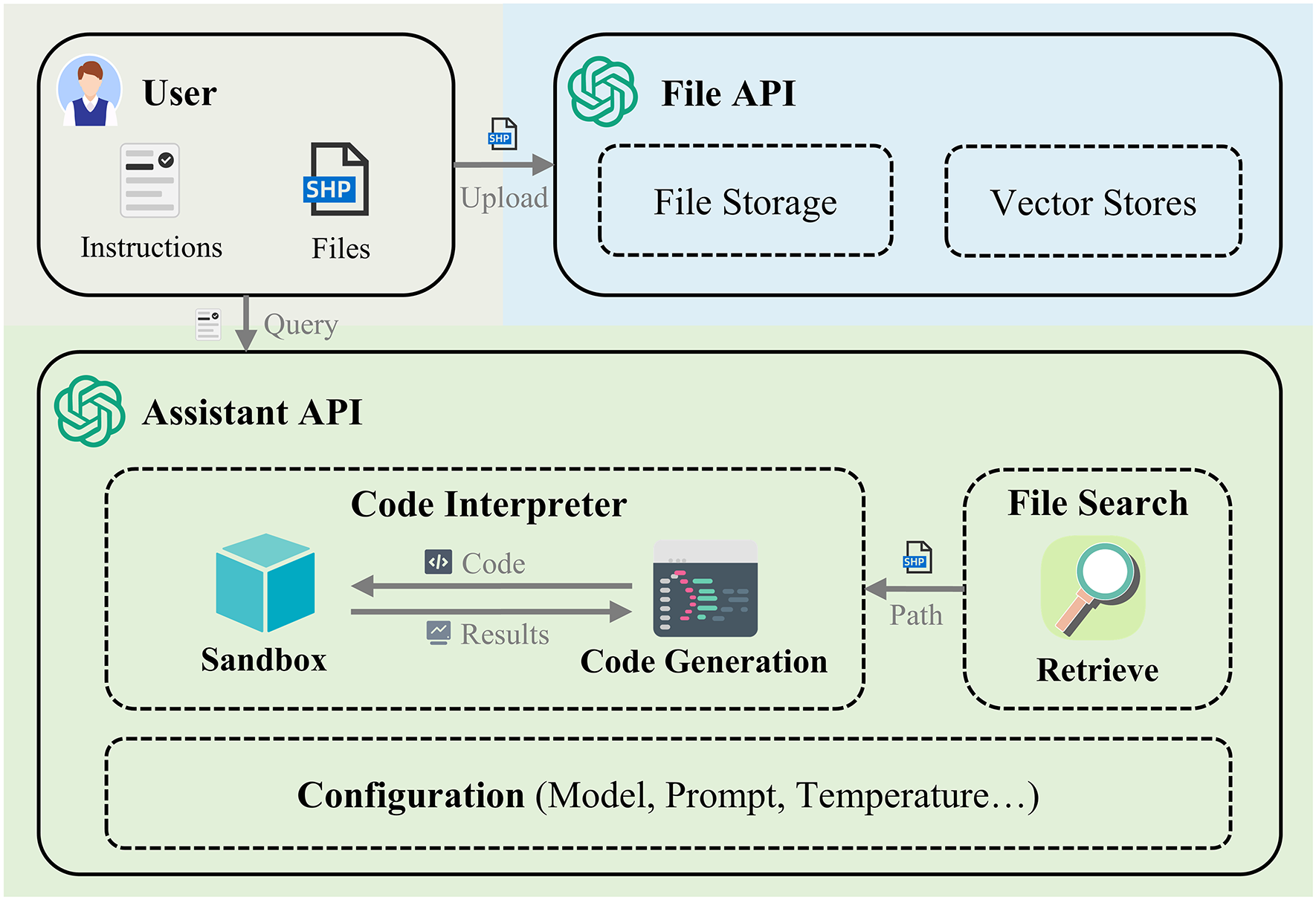}
  \caption{Workflow of how Shapefile tasks are executed using OpenAI's assistant and file APIs.}
  \label{fig:experiment:openai_api}
\end{figure}

\noindent\textbf{Baseline Models}
We compared ShapefileGPT with several models, including GPT-4-Turbo-2024-04-09, GPT-4o-Mini-2024-07-18, and GPT-4o-2024-05-13.
As shown in Fig.~\ref{fig:experiment:openai_api}, we used OpenAI’s assistant and file APIs to evaluate the ability of these GPT models to perform Shapefile tasks.
The process begins with users uploading the relevant Shapefile task files to OpenAI's file storage via the file interface.
The assistant is configured with the necessary parameters, and task instructions are provided.
The assistant accesses the files from the file storage space, generates the required code, and executes it in a sandbox environment to complete the Shapefile task.
Upon completion, the assistant generates a response for the user and stores the result in the file storage space for retrieval.

\begin{figure}
  \centering
  \includegraphics[width=\linewidth]{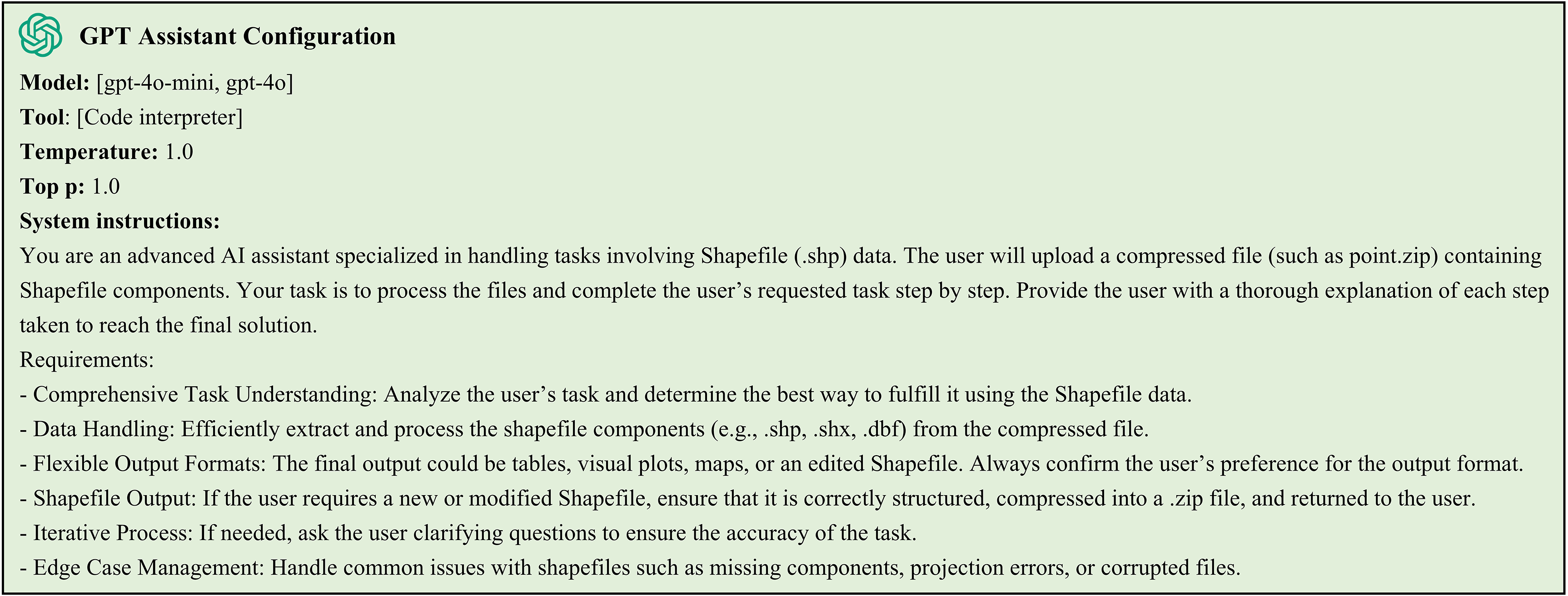}
  \caption{Configuration of GPT assistant used to handle Shapefile tasks.}
  \label{fig:experiment:assistant_configuration}
\end{figure}

The detailed configuration of the assistant is shown in Fig.~\ref{fig:experiment:assistant_configuration}.
Since OpenAI’s file interface does not currently support the direct storage of DBF files from Shapefiles, users must compress the Shapefile data before uploading it to the file storage space.

\noindent\textbf{Evaluation Metrics}
We use accuracy and success rate as the primary metrics to evaluate the performance of baseline models and ShapefileGPT in executing Shapefile tasks.
A task is considered successful if the output file meets the task's requirements.
\textbf{Accuracy} measures the proportion of tasks that the model completes successfully without exceptions.
In ShapefileGPT, exceptions include incorrect or redundant function calls, whereas in baseline models, exceptions refer to the generation of erroneous code.
\textbf{Success rate} measures the proportion of tasks completed successfully, even when exceptions occur during execution.
 
Additionally, based on the Berkeley function calling leaderboard standards \cite{patil2023gorillalargelanguagemodel}, we introduced the metrics of parameter accuracy and parameter repetition rate to further assess ShapefileGPT's performance in function call tasks.
\textbf{Parameter accuracy} measures the proportion of valid parameter types and counts in the function calls generated by the model.
\textbf{Parameter repetition rate} assesses the proportion of redundant function calls generated by the model, reflecting its efficiency in executing tasks.

\subsection{Comparing Task Execution Ability with Different LLMs}
\label{sec:experiment:comparison_with_gpt}
We conducted performance tests on both the baseline Models and ShapefileGPT using the same dataset and performed a detailed analysis of the results.
For the GPT series models, including GPT-4-Turbo-2024-04-09, GPT-4o-Mini-2024-07-18, and GPT-4o-2024-05-13, we recorded the code generation outputs, user responses, and final result files from their task executions.
This data was then used for subsequent analysis to evaluate the performance of these models in executing real-world tasks.

\begin{figure}[h]
  \centering
  \includegraphics[width=0.7\linewidth]{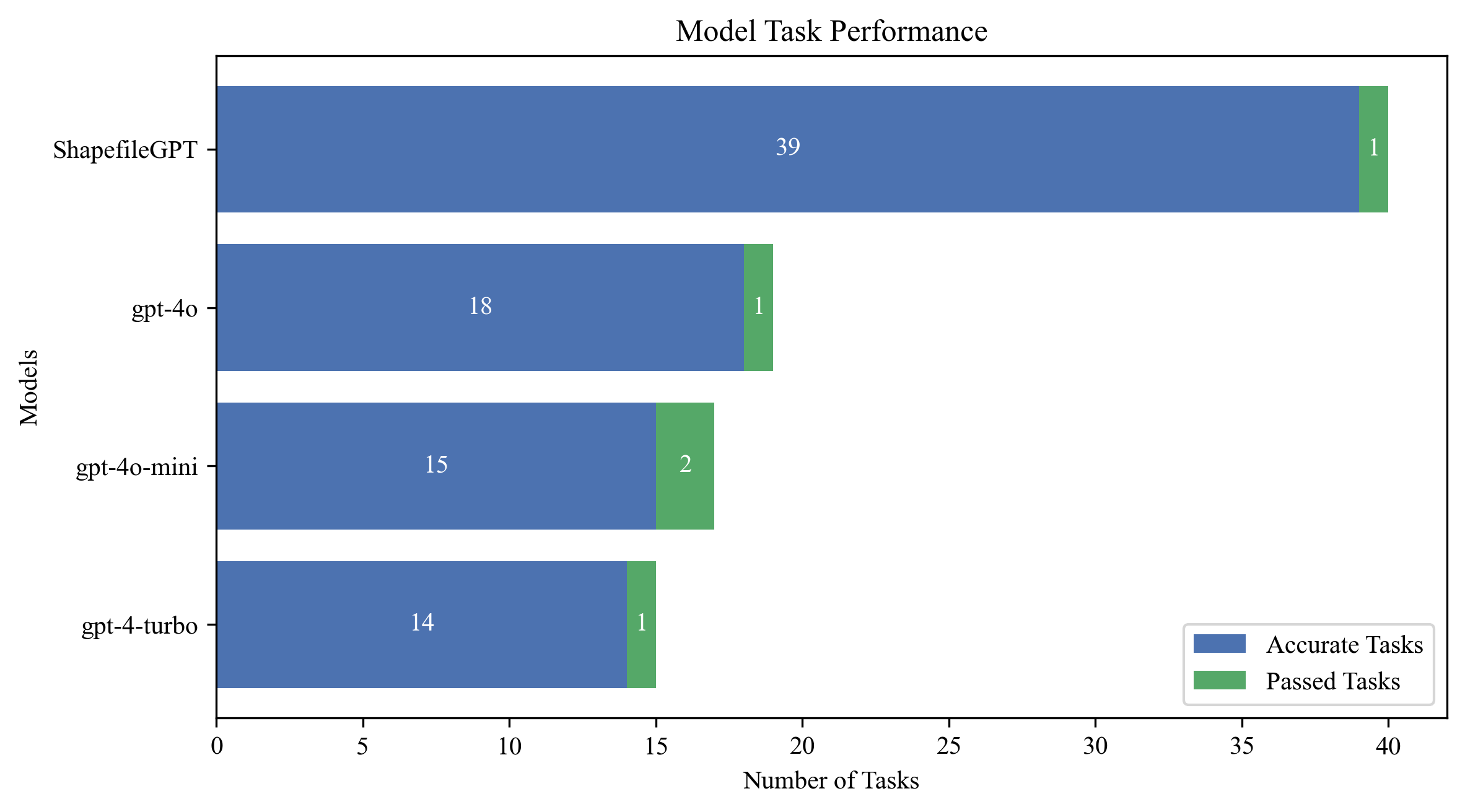}
  \caption{Model task performance comparison between ShapefileGPT and GPT models.}
  \label{fig:experiment:model_task_performance}
\end{figure}

For ShapefileGPT, we manually inspected the function calling records and final outputs for each task to determine whether it accurately completed the task using the same user prompts.
For the GPT models, we reviewed their responses, generated code, and final outputs to assess whether they successfully executed the tasks.
The number of successfully executed tasks for both models on the dataset is shown in Fig.~\ref{fig:experiment:model_task_performance}.

As shown in the experimental results in Table~\ref{tab:experiment:comparison_with_gpt}, GPT-4-Turbo-2024-04-09 achieved an accuracy rate of 33.33\% and a success rate of 35.71\%.
The newer GPT-4o-2024-05-13 performed slightly better, with an accuracy rate of 42.86\% and a success rate of 45.24\%.
In contrast, ShapefileGPT demonstrated significantly superior performance on the same tasks, with an accuracy rate of 92.86\% and a success rate of 95.24\%, far surpassing the GPT models, particularly in its reliability when handling complex GIS tasks.

\begin{table}[h]
\centering
\caption{Performance comparison of ShapefileGPT and GPT models.}
\label{tab:experiment:comparison_with_gpt}
\begin{tabular}{c|ccc}
\toprule
\textbf{Models} & \textbf{Accuracy} & \textbf{Success Rate}\\
\midrule %
GPT-4-Turbo-2024-04-09 & 33.33\% & 35.71\% \\
GPT-4o-Mini-2024-07-18 & 35.71\% & 40.48\% \\
GPT-4o-2024-05-13 & 42.86\% & 45.24\% \\
ShapefileGPT & \textbf{92.86\%} & \textbf{95.24\%} \\
\bottomrule 
\end{tabular}
\end{table}

\subsection{Comparing Different Configurations of ShapefileGPT}
\label{sec:experiment:comparison_with_different_config}
In this section, we conducted performance comparison experiments with multiple model configurations of ShapefileGPT to assess how foundational models perform in task environments.
By testing various model combinations, we aim to identify which configurations excel in the system’s agent tasks, thereby helping to establish ShapefileGPT's performance limits.

In the experimental design, we defined two key modules of the ShapefileGPT system: the planner, responsible for task decomposition and instruction generation, and the worker, responsible for executing specific operations.
To ensure that the planner module efficiently breaks down complex tasks and generates reasonable execution instructions, the foundational model selected must possess strong reasoning and planning abilities.
In contrast, the worker module focuses on selecting the appropriate API from the contextual API documentation to execute specific tasks.
Therefore, the model it relies on must possess strong capabilities in retaining contextual information and accurately calling APIs, ensuring that no important document content is overlooked.

\begin{table}
\centering

\caption{Performance of different ShapefileGPT configurations.}
\label{tab:experiment:comparison_with_different_config}

\resizebox{\linewidth}{!}{
\begin{tabular}{c|cc|ccc}
\toprule

\textbf{ID} &\textbf{Planner Models} & \textbf{Worker Models} & \textbf{Accuracy} & \textbf{Success Rate} & \textbf{Calls Repetition Rate} \\

\midrule %
1 & GPT-4o-2024-05-13 & GPT-4o-Mini-2024-07-18 & \textbf{92.86\%} & \textbf{95.24}\%  & 0.1960\\
2 &GPT-4o-Mini-2024-07-18 & GPT-4o-Mini-2024-07-18 & 90.48\% & \textbf{95.24}\%  & \textbf{0.0079}\\
3 &GPT-4o-Mini-2024-07-18 & GPT-3.5-Turbo-0125 & 7.14\% & 23.81\%  & 1.5543\\

\bottomrule
\end{tabular}
}
\end{table}







As shown in the experimental results in Table~\ref{tab:experiment:comparison_with_different_config}, ShapefileGPT under configuration 1 achieved the highest accuracy and success rates, at 92.86\% and 95.24\%, respectively.
In terms of the repeat call rate, configuration 2 demonstrated a significantly lower rate of 0.0079 compared to other model combinations, indicating more efficient task completion with fewer redundant function calls.
These results suggest that the GPT-4o series models not only excel at handling complex tasks but also significantly reduce system resource waste.

The accuracy of parameter calls is a key metric for evaluating a model's ability to generate valid function calls.
In all configurations, this value was 100\%, indicating that the GPT models successfully generated correct function calls based on standardized API documentation.
This metric validates the reasoning capabilities of the GPT models, demonstrating their proficiency in generating accurate function calls, which lays the groundwork for more advanced tool utilization.

When the worker model was replaced with GPT-3.5-Turbo (configurations 4 and 5), both the accuracy and success rate dropped significantly, with the accuracy at 7.14\% and the success rate at 23.81\%, while the repeat call rate increased to 1.5543.
The primary cause of this performance decline was GPT-3.5-Turbo's limited function calling ability.
It failed to effectively select the appropriate APIs from the documentation to execute tasks assigned by the planner, leading to the task stalls.
This inefficiency forced the planner to issue the same instructions repeatedly, significantly reducing the system’s overall efficiency.
These results demonstrate that GPT-3.5-Turbo lags considerably behind the more advanced GPT-4o series in function calling capabilities and is unable to effectively support the accurate functioning of the ShapefileGPT system.

\subsection{Validation of the Agent Module Through Case Studies}
\label{sec:experiment:module_validation}
This section validated the effectiveness of the planner and worker modules through multiple GIS case studies, focusing on the planner’s ability to interpret user instructions and decompose tasks, as well as the worker module’s precision in calling GIS functions and executing them.

\subsubsection{Case 1: Spatial Allocation of Points by Distance}
\label{sec:experiment:case1}
This case demonstrates how the planner and worker modules perform a distance-based spatial allocation of point features.
The user provided a point feature dataset and requested the generation of voronoi polygons, along with a 500-meter buffer analysis around the point features.
The planner module decomposed the task into three steps: (1) generate Voronoi polygons, (2) create a 500-meter buffer, and (3) clip the buffer using the voronoi polygons.
The worker module accurately executed these tasks by sequentially invoking the appropriate GIS functions.

\begin{figure}[h]
  \centering
  \includegraphics[width=\linewidth]{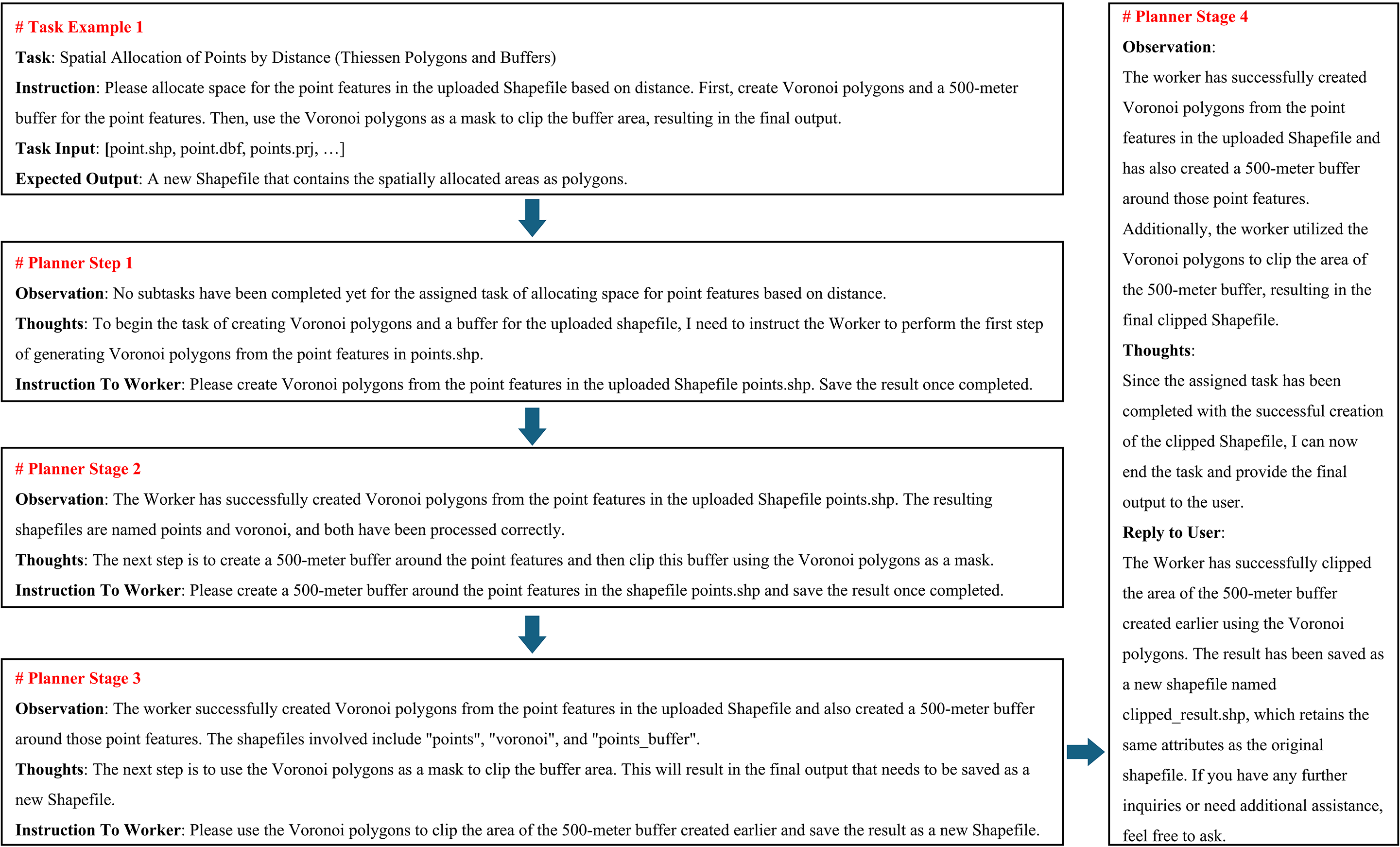}
  \caption{The step-by-step task decomposition by the planner module for spatial allocation of points.}
  \label{fig:experiment:task1_planner}
\end{figure}
\begin{figure}[H]
  \centering
  \includegraphics[width=\linewidth]{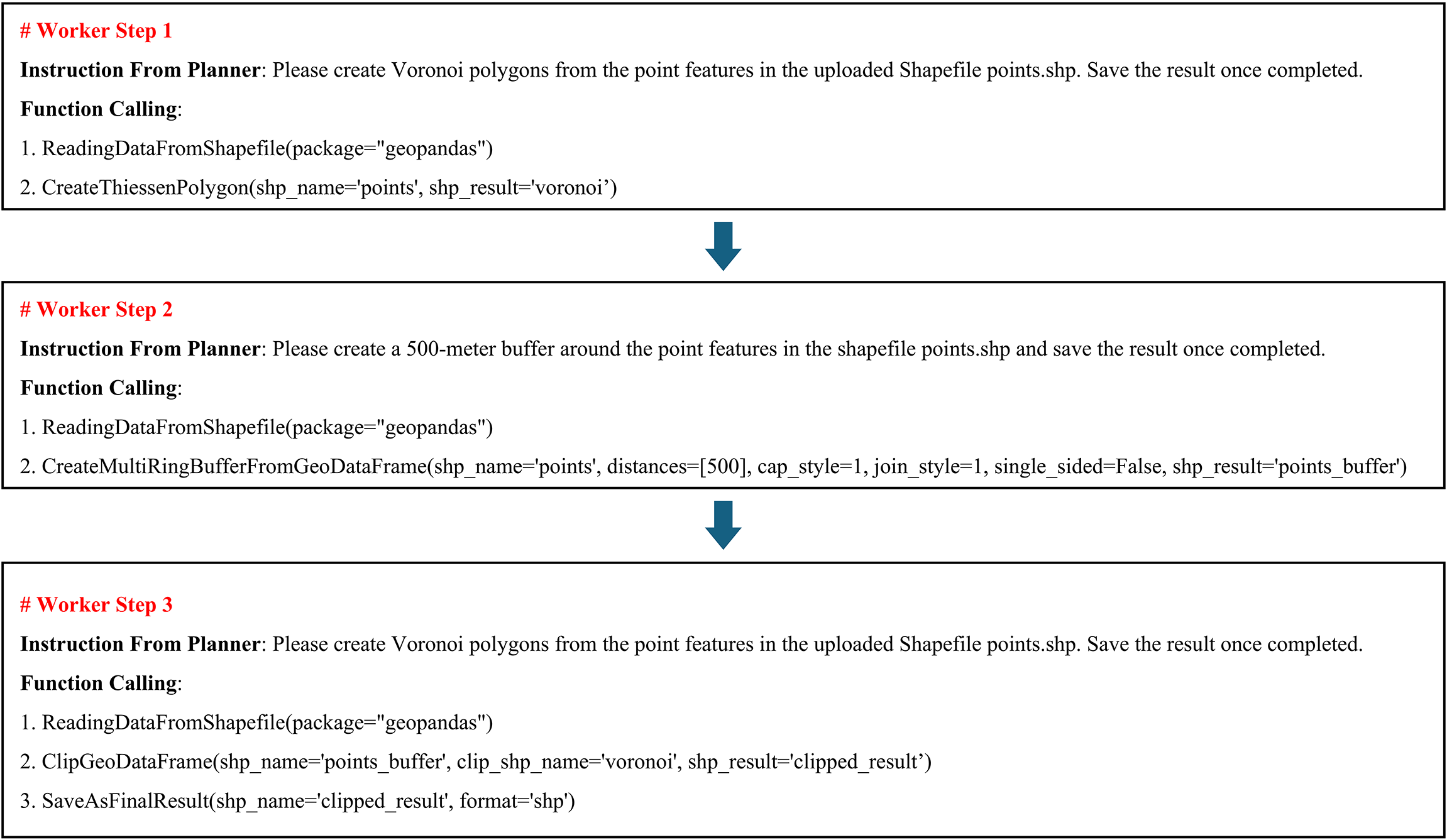}
  \caption{The stepwise execution by the worker module for spatial allocation of points.}
  \label{fig:experiment:task1_worker}
\end{figure}

Fig.~\ref{fig:experiment:task1_planner} and Fig.~\ref{fig:experiment:task1_worker} display the outputs from the planner and worker at each stage.
The results demonstrate that the planner effectively decomposes subtasks, while the worker successfully executes complex GIS tasks and generates the expected spatial data outputs.
These findings confirm the robustness and reliability of the system.

\subsubsection{Case 2: Analyze Disaster Impact Buffers}
\label{sec:experiment:case2}
In this case, we demonstrate how the planner and worker modules perform a disaster impact buffer analysis.
The task involves creating multiple concentric buffers around the disaster area to analyze potential impact zones, and performing spatial analysis on the roads within the affected area.

\begin{figure}[h]
  \centering
  \includegraphics[width=\linewidth]{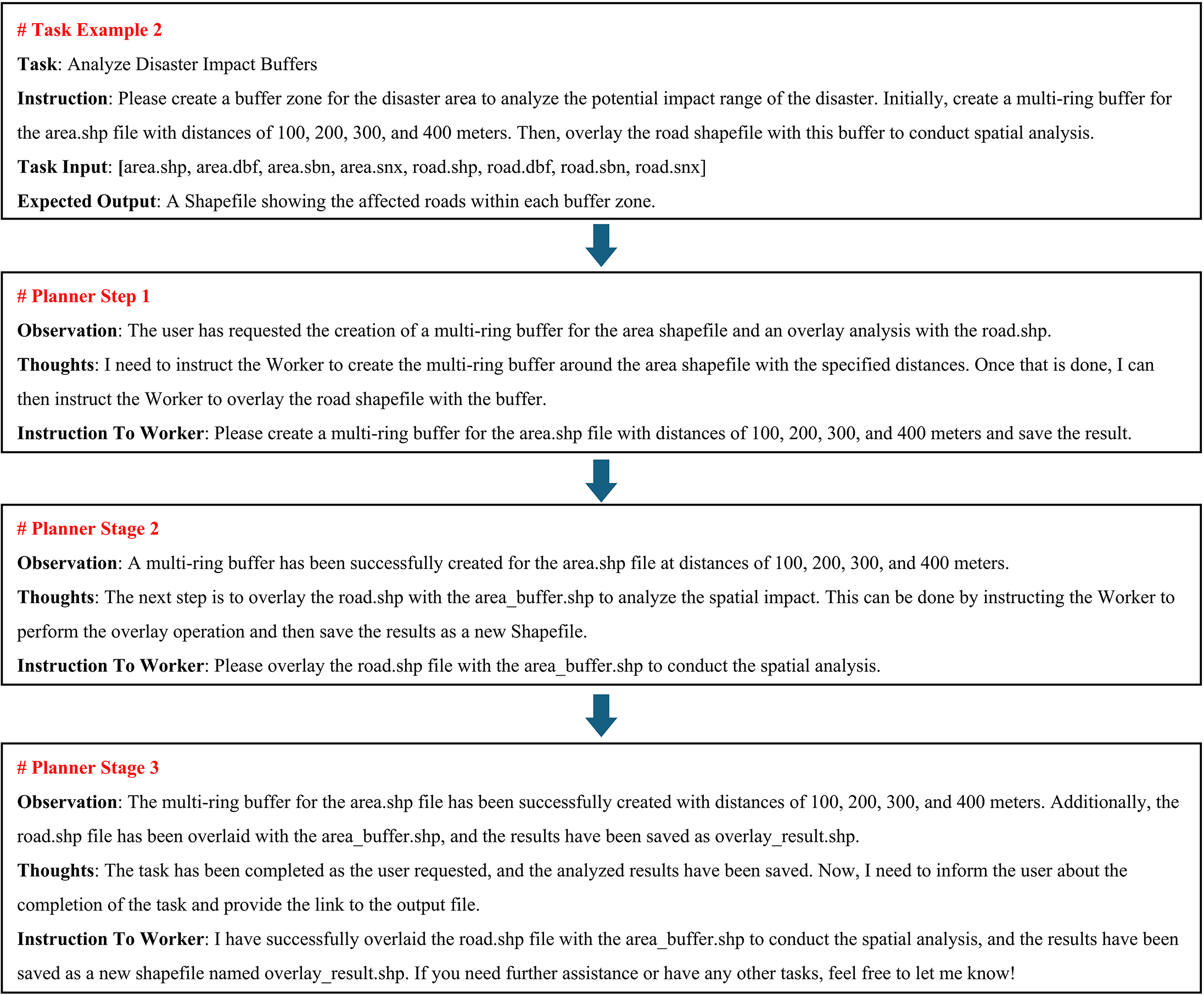}
  \caption{The step-by-step task decomposition by the planner module for disaster impact buffer analysis.}
  \label{fig:experiment:task2_planner}
\end{figure}
\begin{figure}[h!]
  \centering
  \includegraphics[width=\linewidth]{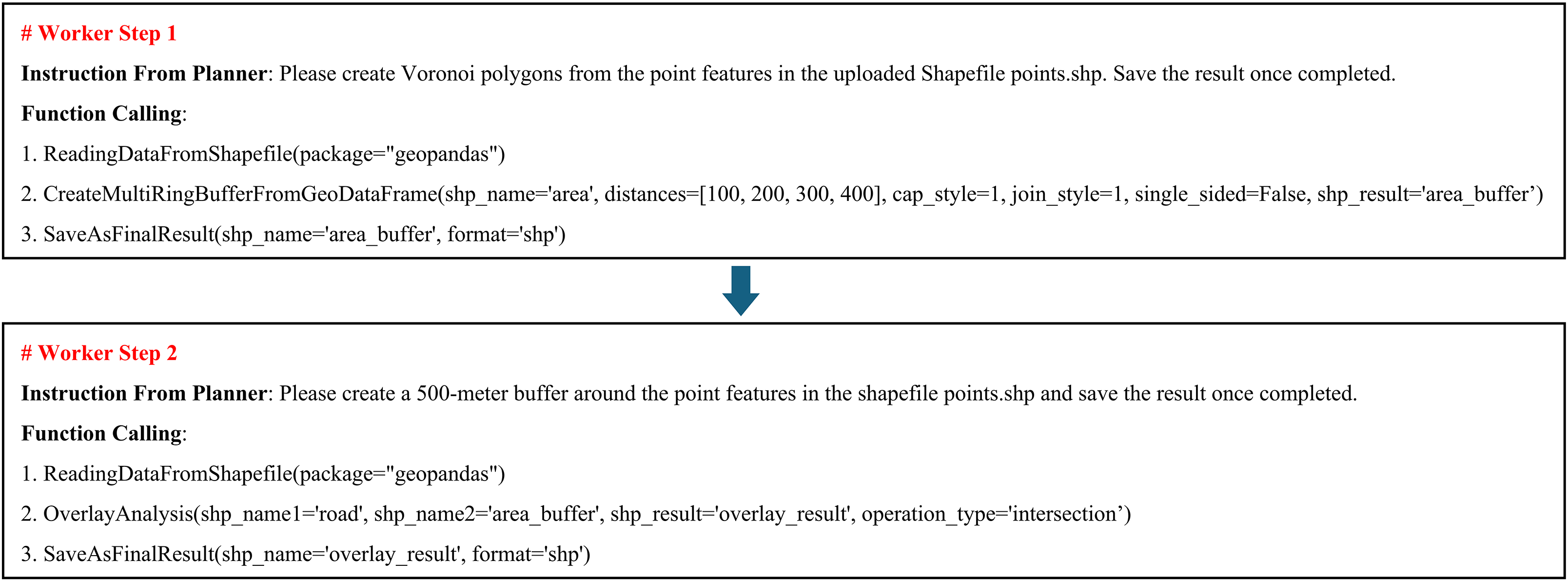}
  \caption{The stepwise execution by the worker module for disaster impact buffer analysis.}
  \label{fig:experiment:task2_worker}
\end{figure}

Fig.~\ref{fig:experiment:task2_planner} and Fig.~\ref{fig:experiment:task2_worker} display the outputs from the planner and worker at each stage.
Specifically, the planner module divided the task into two main steps based on the task requirements: first, generating multiple concentric buffers around the area file at specified distances, and second, overlaying the generated buffers with the road Shapefile for spatial analysis.
The worker module accurately invoked the necessary GIS functions, generating the buffers, overlaying them with the road data, and outputting information on the roads within the disaster impact zones.

\subsection{Ablation Study of Planner Agent}
\label{sec:experiment:ablation_planner}

\begin{table}[h]
\centering
\caption{Ablation study results for the planner module in ShapefileGPT.}
\label{tab:experiment:ablation_planner}

\resizebox{\linewidth}{!}{
\begin{tabular}{c|cc|ccc}
\toprule

\textbf{ID} &\textbf{Planner Models} & \textbf{Worker Models} & \textbf{Accuracy} & \textbf{Success Rate} & \textbf{Calls Repetition Rate} \\

\midrule %
1 & GPT-4o-2024-05-13 & GPT-4o-Mini-2024-07-18 & \textbf{92.86\%} & \textbf{95.24}\% & 0.1960\\
2 &GPT-4o-Mini-2024-07-18 & GPT-4o-Mini-2024-07-18 & 90.48\% & \textbf{95.24}\% & \textbf{0.0079}\\
3 &\multicolumn{1}{c}{\ding{55}} & GPT-4o-Mini-2024-07-18 & 88.06\% & 92.86\% & 0.0566\\
4 &GPT-4o-Mini-2024-07-18 & GPT-3.5-Turbo-0125 & 7.14\% & 23.81\% & 1.5543\\
5 &\multicolumn{1}{c}{\ding{55}} & GPT-3.5-Turbo-0125 & 11.94\% & 19.05\% & 0.2274\\

\bottomrule
\end{tabular}
}
\end{table}






In this section, we conduct ablation experiments on the planner agent module to explore its role and impact within ShapefileGPT.
The experiment is designed to assess the planner's performance in guiding the worker module, with a particular focus on changes in success rate and task execution efficiency.
Table~\ref{tab:experiment:ablation_planner} summarizes the experimental results across various configurations.

The experimental results indicate that when both the planner and worker are configured with the GPT-4o-Mini-2024-07-18 model (configuration 1), the system’s overall performance is strong, achieving the accuracy and success rates of 90.48\% and 95.24\%, respectively.
However, when the planner is removed from configuration 1, the accuracy and success rates drop to 88.06\% and 92.86\%.
These results demonstrate that the planner enhances the system's ability to complete Shapefile tasks by efficiently allocating tasks and providing guidance to the worker, ensuring proper task execution.

However, performance drops significantly when the worker model is replaced with GPT-3.5-Turbo-0125, as seen in configurations 3 and 4.
In configuration 3, although the worker’s accuracy is only 7.14\%, the planner raises the success rate to 23.81\% by issuing repeated instructions.
This suggests that when the worker fails to execute tasks, the planner automatically detects the failure and guides the worker to retry.
This is further supported by the increased call repetition rate of 1.5543.
The increase in the repetition rate indicates that the planner detects the worker’s errors and sends repeated instructions to ensure task completion, though this increases the number of function calls, negatively affecting overall task efficiency.

When the system operates without the planner (configuration 4), its accuracy and success rates fall to 11.94\% and 19.05\%, respectively, with the repetition rate dropping to 0.2274.
This indicates that without the planner, although the worker can occasionally complete tasks, the system’s overall fault tolerance is reduced.
Without the ability to detect errors or initiate retries, the task completion rate declines.

Based on the analysis of the experimental results, the role of the planner in the ShapefileGPT system is evident in several key areas.
First, the planner enhances the system’s task success rate by effectively decomposing complex tasks and guiding the worker module.
This is especially beneficial when worker performance is suboptimal, as the planner compensates for worker deficiencies and improves overall efficiency.
Furthermore, the planner has error detection and task retry capabilities.
If the worker makes an error during task execution, the planner detects it and issues multiple instructions to retry the task, ensuring successful completion.
Although this process increases the number of function calls, it significantly improves the system’s fault tolerance and task success rate, particularly in handling complex tasks.
Lastly, as the system’s central coordinator, the planner plays a crucial role in task allocation and oversight.
It allocates tasks based on complexity and intervenes when failures occur, ensuring system stability and reliability.

\begin{table}[h]
\centering

\caption{Ablation study of task examples and API examples for worker.}
\label{tab:experiment:ablation_fewshot}

\resizebox{0.7\textwidth}{!}{
\begin{tabular}{c|cc|ccc}
\toprule
\textbf{ID} & \textbf{Task Example} & \textbf{API Example} & \textbf{Accuracy} & \textbf{Sucess Rate} & \textbf{Calls Repetition Rate} \\
\midrule
1 & & & 71.43\% & 78.57\% & 0.1278 \\
2 & \ding{51} & & 78.57\% & 80.95\% & 0.0931 \\
3 & & \ding{51} & 85.71\% & 88.10\% & 0.0897 \\
4 & \ding{51} & \ding{51} & 88.10\% & 92.86\% & 0.0566 \\
\bottomrule
\end{tabular}
}
\end{table}




\subsection{Ablation Study of Worker Few Shot Prompting}
\label{sec:experiment:ablation_fewshot}
We conducted ablation studies on the worker prompts in ShapefileGPT, specifically examining how the task example in the worker system prompts and the API example in the documentation influence task execution performance.
In this experiment, the task example refers to a pre-input few-shot task example embedded in the worker system prompts, similar to a preloaded conversation.
The API example refers to few-shot examples provided for each function in the API documentation, designed to guide the worker in generating accurate function calls.

In the experiment, we sequentially removed the task example from the worker system prompts and the API example from the documentation, then tested the entire dataset.
The experimental results, presented in Table~\ref{tab:experiment:ablation_fewshot}, show that as the prompts are removed, the model's overall performance declines in terms of accuracy and success rate, while the repetition rate of function calls increases.

Specifically, when the task example was removed, worker accuracy dropped from 88.10\% to 71.43\%, and the success rate dropped from 92.86\% to 78.57\%.
Similarly, when the API example was removed, worker accuracy dropped to 85.71\%, and the success rate fell to 88.10\%.
Notably, across all experiments, the function call parameter accuracy remained at 100\%, indicating that even with fewer prompts, the worker’s understanding and generation of function parameters remained consistent.
However, with fewer prompts, the function call repetition rate increased, particularly when all prompts were removed, rising to 0.1278 compared to 0.0566 with full prompts.

Overall, these findings indicate that the task example and API example prompts significantly enhance the worker’s task execution and function call efficiency, while their absence leads to a notable decline in system performance.
The pre-provided task and function call examples play a crucial role in guiding the worker’s reasoning, reducing unnecessary function call repetitions, and ultimately improving overall system performance.
These results offer valuable insights for designing prompts in future iterations.
\section{Discussion}
\subsection{Strengths and Limitations of Multi-Agent Architecture}
\textbf{Specialization and Modularity}
The key advantage of a multi-agent system is its ability to divide tasks and promote collaboration.
By dividing labor, each agent can focus on specific subtasks, which enhancing overall task efficiency and quality.
In ShapefileGPT, the planner is solely responsible for task allocation, without concern for execution details, while the worker focuses on task completion.
This separation of roles reduces the cognitive load on each agent, optimizing their performance.

\textbf{Fault Tolerance and Error Recovery}
In complex GIS operations, where errors are likely, the ability to handle them is critical.
Multi-agent systems demonstrate greater adaptability in challenging environments.
As shown in Table 5, ShapefileGPT configurations that include a planner achieve higher success rates.
In ShapefileGPT, the planner detects worker failures, retries tasks, and adjusts the order of subtasks as needed.
This self-correction mechanism enables complex tasks to progress without the need for continuous manual intervention.

However, the limitations of the multi-agent architecture manifest in the following ways.

\textbf{Increased Computational Overhead}
In a multi-agent system, communication and coordination between agents introduce additional overhead, as observed in ShapefileGPT.
During multi-step GIS operations, the planner interacts with the worker multiple times, leading to delays and increased token usage.

\textbf{Complexity in Error Management}
Although the planner incorporates error detection and recovery mechanisms, more complex error scenarios can arise.
In such cases, despite the planner’s best efforts to retry, the worker may repeatedly fail to complete tasks.
This is evident in configurations 3 and 4 in Table 5.
After replacing the worker model with GPT-3.5-Turbo-0125, task success rates dropped significantly in both configurations due to reduced function-calling capabilities.
Configuration 3 exhibited a higher repetition rate than configuration 4, indicating that the planner attempted multiple retries.

\subsection{Limitations and Future Improvements}
\textbf{Hallucinations and Randomness in LLMs}
Despite their powerful reasoning and generation capabilities, large language models can exhibit hallucinations when handling complex tasks, producing inaccurate or irrelevant information.
This issue is particularly pronounced in Shapefile tasks, where geometric and spatial analysis require precise operations and outcomes.
Additionally, LLMs' tendency to generate varying outputs under identical configurations introduces uncertainty, potentially leading to inconsistencies in automated workflows.
This variability increases the need for result validation and presents risks in automating Shapefile processing.
Although errors are minimized through supervision and correction within the multi-agent architecture, hallucinations can still affect result accuracy, particularly when processing complex boundaries or irregular data.
Future improvements could focus on optimizing the model’s context-handling mechanisms to mitigate hallucinations.

\textbf{Token Consumption}
ShapefileGPT's operation relies on large language models through API calls, which is efficient but leads to high token consumption, particularly for complex tasks.
This results in increased computational costs and financial overhead.
As the application scales to handle larger or more complex GIS tasks, these cost issue will become more pronounced.
To mitigate these challenges, future strategies could involve incorporating local model inference to optimize token usage and reduce computational expenses, thereby enhancing ShapefileGPT's cost-effectiveness.

\textbf{Dataset Size}
For this evaluation, we used a relatively small Shapefile task dataset, focusing on a limited range of geometric operations and spatial queries.
This limitation restricted our ability to comprehensively assess the model's performance on more complex and diverse vector data tasks.
While ShapefileGPT performed well on the current dataset, this performance does not fully reflect its applicability to more complex and diverse tasks.
Future work should expand the dataset to include additional tasks and categories, covering a broader range of operations and query scenarios, to enable a more comprehensive evaluation of the model’s performance across diverse applications.
\section{Conclusion}
We proposed ShapefileGPT, a multi-agent framework that enables users to interact with the system using natural language.
This framework automatically decomposes Shapefile tasks proposed by users into subtasks and completes them through the collaboration of multiple agents.
To evaluate the performance of our agents, we created a dataset encompassing multiple task categories, covering common vector data spatial analysis operations.
These categories include geometric queries, spatial overlay, buffer analysis, and more, representing typical scenarios in real-world GIS applications.

The experimental results demonstrate that ShapefileGPT completes tasks with high accuracy and effectively utilizes the specialized vector data analysis function modules we designed, ensuring the successful execution of complex spatial analysis tasks.
Case studies show that in the multi-agent architecture, the planner agent is responsible for task decomposition and planning, while the worker agent handles specific execution.
Together, they collaborate efficiently to complete the Shapefile tasks proposed by users.
Ablation experiments further validated the independent contributions and importance of each agent module in task execution.

ShapefileGPT not only provides GIS professionals with an efficient automation tool but also significantly lowers the technical barriers for researchers in non-GIS fields to process spatial data, fostering interdisciplinary collaboration.
The design of this versatile framework highlights the broad potential of large language models in handling complex geospatial tasks, offering valuable insights for future agent development in the GIS domain.


\bibliography{references}

\begin{thebibliography}{10}

\bibitem{wang2024researchspatialdataintelligent}
Shaohua Wang, Xing Xie, Yong Li, Danhuai Guo, Zhi Cai, Yu~Liu, Yang Yue, Xiao Pan, Feng Lu, Huayi Wu, Zhipeng Gui, Zhiming Ding, Bolong Zheng, Fuzheng Zhang, Jingyuan Wang, Zhengchao Chen, Hao Lu, Jiayi Li, Peng Yue, Wenhao Yu, Yao Yao, Leilei Sun, Yong Zhang, Longbiao Chen, Xiaoping Du, Xiang Li, Xueying Zhang, Kun Qin, Zhaoya Gong, Weihua Dong, and Xiaofeng Meng.
\newblock Research on the spatial data intelligent foundation model, 2024.

\bibitem{temporary-citekey-5333}
Michael~John De~Smith, Michael~F Goodchild, and Paul Longley.
\newblock {\em {Geospatial analysis: a comprehensive guide to principles, techniques and software tools}}.
\newblock Troubador publishing ltd, 2007.

\bibitem{openai2024gpt4technicalreport}
OpenAI, Josh Achiam, Steven Adler, Sandhini Agarwal, Lama Ahmad, Ilge Akkaya, Florencia~Leoni Aleman, Diogo Almeida, Janko Altenschmidt, Sam Altman, Shyamal Anadkat, Red Avila, Igor Babuschkin, Suchir Balaji, Valerie Balcom, Paul Baltescu, Haiming Bao, Mohammad Bavarian, Jeff Belgum, Irwan Bello, Jake Berdine, Gabriel Bernadett-Shapiro, Christopher Berner, Lenny Bogdonoff, Oleg Boiko, Madelaine Boyd, Anna-Luisa Brakman, Greg Brockman, Tim Brooks, Miles Brundage, Kevin Button, Trevor Cai, Rosie Campbell, Andrew Cann, Brittany Carey, Chelsea Carlson, Rory Carmichael, Brooke Chan, Che Chang, Fotis Chantzis, Derek Chen, Sully Chen, Ruby Chen, Jason Chen, Mark Chen, Ben Chess, Chester Cho, Casey Chu, Hyung~Won Chung, Dave Cummings, Jeremiah Currier, Yunxing Dai, Cory Decareaux, Thomas Degry, Noah Deutsch, Damien Deville, Arka Dhar, David Dohan, Steve Dowling, Sheila Dunning, Adrien Ecoffet, Atty Eleti, Tyna Eloundou, David Farhi, Liam Fedus, Niko Felix, Simón~Posada Fishman, Juston Forte, Isabella Fulford, Leo
  Gao, Elie Georges, Christian Gibson, Vik Goel, Tarun Gogineni, Gabriel Goh, Rapha Gontijo-Lopes, Jonathan Gordon, Morgan Grafstein, Scott Gray, Ryan Greene, Joshua Gross, Shixiang~Shane Gu, Yufei Guo, Chris Hallacy, Jesse Han, Jeff Harris, Yuchen He, Mike Heaton, Johannes Heidecke, Chris Hesse, Alan Hickey, Wade Hickey, Peter Hoeschele, Brandon Houghton, Kenny Hsu, Shengli Hu, Xin Hu, Joost Huizinga, Shantanu Jain, Shawn Jain, Joanne Jang, Angela Jiang, Roger Jiang, Haozhun Jin, Denny Jin, Shino Jomoto, Billie Jonn, Heewoo Jun, Tomer Kaftan, Łukasz Kaiser, Ali Kamali, Ingmar Kanitscheider, Nitish~Shirish Keskar, Tabarak Khan, Logan Kilpatrick, Jong~Wook Kim, Christina Kim, Yongjik Kim, Jan~Hendrik Kirchner, Jamie Kiros, Matt Knight, Daniel Kokotajlo, Łukasz Kondraciuk, Andrew Kondrich, Aris Konstantinidis, Kyle Kosic, Gretchen Krueger, Vishal Kuo, Michael Lampe, Ikai Lan, Teddy Lee, Jan Leike, Jade Leung, Daniel Levy, Chak~Ming Li, Rachel Lim, Molly Lin, Stephanie Lin, Mateusz Litwin, Theresa Lopez, Ryan
  Lowe, Patricia Lue, Anna Makanju, Kim Malfacini, Sam Manning, Todor Markov, Yaniv Markovski, Bianca Martin, Katie Mayer, Andrew Mayne, Bob McGrew, Scott~Mayer McKinney, Christine McLeavey, Paul McMillan, Jake McNeil, David Medina, Aalok Mehta, Jacob Menick, Luke Metz, Andrey Mishchenko, Pamela Mishkin, Vinnie Monaco, Evan Morikawa, Daniel Mossing, Tong Mu, Mira Murati, Oleg Murk, David Mély, Ashvin Nair, Reiichiro Nakano, Rajeev Nayak, Arvind Neelakantan, Richard Ngo, Hyeonwoo Noh, Long Ouyang, Cullen O'Keefe, Jakub Pachocki, Alex Paino, Joe Palermo, Ashley Pantuliano, Giambattista Parascandolo, Joel Parish, Emy Parparita, Alex Passos, Mikhail Pavlov, Andrew Peng, Adam Perelman, Filipe de~Avila Belbute~Peres, Michael Petrov, Henrique~Ponde de~Oliveira~Pinto, Michael, Pokorny, Michelle Pokrass, Vitchyr~H. Pong, Tolly Powell, Alethea Power, Boris Power, Elizabeth Proehl, Raul Puri, Alec Radford, Jack Rae, Aditya Ramesh, Cameron Raymond, Francis Real, Kendra Rimbach, Carl Ross, Bob Rotsted, Henri Roussez,
  Nick Ryder, Mario Saltarelli, Ted Sanders, Shibani Santurkar, Girish Sastry, Heather Schmidt, David Schnurr, John Schulman, Daniel Selsam, Kyla Sheppard, Toki Sherbakov, Jessica Shieh, Sarah Shoker, Pranav Shyam, Szymon Sidor, Eric Sigler, Maddie Simens, Jordan Sitkin, Katarina Slama, Ian Sohl, Benjamin Sokolowsky, Yang Song, Natalie Staudacher, Felipe~Petroski Such, Natalie Summers, Ilya Sutskever, Jie Tang, Nikolas Tezak, Madeleine~B. Thompson, Phil Tillet, Amin Tootoonchian, Elizabeth Tseng, Preston Tuggle, Nick Turley, Jerry Tworek, Juan Felipe~Cerón Uribe, Andrea Vallone, Arun Vijayvergiya, Chelsea Voss, Carroll Wainwright, Justin~Jay Wang, Alvin Wang, Ben Wang, Jonathan Ward, Jason Wei, CJ~Weinmann, Akila Welihinda, Peter Welinder, Jiayi Weng, Lilian Weng, Matt Wiethoff, Dave Willner, Clemens Winter, Samuel Wolrich, Hannah Wong, Lauren Workman, Sherwin Wu, Jeff Wu, Michael Wu, Kai Xiao, Tao Xu, Sarah Yoo, Kevin Yu, Qiming Yuan, Wojciech Zaremba, Rowan Zellers, Chong Zhang, Marvin Zhang, Shengjia
  Zhao, Tianhao Zheng, Juntang Zhuang, William Zhuk, and Barret Zoph.
\newblock Gpt-4 technical report, 2024.

\bibitem{10.1145/3616855.3635752}
Yuan Sui, Mengyu Zhou, Mingjie Zhou, Shi Han, and Dongmei Zhang.
\newblock Table meets llm: Can large language models understand structured table data? a benchmark and empirical study.
\newblock In {\em Proceedings of the 17th ACM International Conference on Web Search and Data Mining}, WSDM '24, page 645–654, New York, NY, USA, 2024. Association for Computing Machinery.

\bibitem{zha2023tablegptunifyingtablesnature}
Liangyu Zha, Junlin Zhou, Liyao Li, Rui Wang, Qingyi Huang, Saisai Yang, Jing Yuan, Changbao Su, Xiang Li, Aofeng Su, Tao Zhang, Chen Zhou, Kaizhe Shou, Miao Wang, Wufang Zhu, Guoshan Lu, Chao Ye, Yali Ye, Wentao Ye, Yiming Zhang, Xinglong Deng, Jie Xu, Haobo Wang, Gang Chen, and Junbo Zhao.
\newblock Tablegpt: Towards unifying tables, nature language and commands into one gpt, 2023.

\bibitem{xu2024evaluatinglargelanguagemodels}
Liuchang Xu, Shuo Zhao, Qingming Lin, Luyao Chen, Qianqian Luo, Sensen Wu, Xinyue Ye, Hailin Feng, and Zhenhong Du.
\newblock Evaluating large language models on spatial tasks: A multi-task benchmarking study, 2024.

\bibitem{10.1145/3615886.3627745}
Peter Mooney, Wencong Cui, Boyuan Guan, and Levente Juh\'{a}sz.
\newblock Towards understanding the geospatial skills of chatgpt: Taking a geographic information systems (gis) exam.
\newblock In {\em Proceedings of the 6th ACM SIGSPATIAL International Workshop on AI for Geographic Knowledge Discovery}, GeoAI '23, page 85–94, New York, NY, USA, 2023. Association for Computing Machinery.

\bibitem{NEURIPS2023_43e9d647}
Jiawei Liu, Chunqiu~Steven Xia, Yuyao Wang, and LINGMING ZHANG.
\newblock Is your code generated by chatgpt really correct? rigorous evaluation of large language models for code generation.
\newblock In A.~Oh, T.~Naumann, A.~Globerson, K.~Saenko, M.~Hardt, and S.~Levine, editors, {\em Advances in Neural Information Processing Systems}, volume~36, pages 21558--21572. Curran Associates, Inc., 2023.

\bibitem{NEURIPS2022_9d560961}
Jason Wei, Xuezhi Wang, Dale Schuurmans, Maarten Bosma, brian ichter, Fei Xia, Ed~Chi, Quoc~V Le, and Denny Zhou.
\newblock Chain-of-thought prompting elicits reasoning in large language models.
\newblock In S.~Koyejo, S.~Mohamed, A.~Agarwal, D.~Belgrave, K.~Cho, and A.~Oh, editors, {\em Advances in Neural Information Processing Systems}, volume~35, pages 24824--24837. Curran Associates, Inc., 2022.

\bibitem{bai2022traininghelpfulharmlessassistant}
Yuntao Bai, Andy Jones, Kamal Ndousse, Amanda Askell, Anna Chen, Nova DasSarma, Dawn Drain, Stanislav Fort, Deep Ganguli, Tom Henighan, Nicholas Joseph, Saurav Kadavath, Jackson Kernion, Tom Conerly, Sheer El-Showk, Nelson Elhage, Zac Hatfield-Dodds, Danny Hernandez, Tristan Hume, Scott Johnston, Shauna Kravec, Liane Lovitt, Neel Nanda, Catherine Olsson, Dario Amodei, Tom Brown, Jack Clark, Sam McCandlish, Chris Olah, Ben Mann, and Jared Kaplan.
\newblock Training a helpful and harmless assistant with reinforcement learning from human feedback, 2022.

\bibitem{zhong2024evaluationopenaio1opportunities}
Tianyang Zhong, Zhengliang Liu, Yi~Pan, Yutong Zhang, Yifan Zhou, Shizhe Liang, Zihao Wu, Yanjun Lyu, Peng Shu, Xiaowei Yu, Chao Cao, Hanqi Jiang, Hanxu Chen, Yiwei Li, Junhao Chen, Huawen Hu, Yihen Liu, Huaqin Zhao, Shaochen Xu, Haixing Dai, Lin Zhao, Ruidong Zhang, Wei Zhao, Zhenyuan Yang, Jingyuan Chen, Peilong Wang, Wei Ruan, Hui Wang, Huan Zhao, Jing Zhang, Yiming Ren, Shihuan Qin, Tong Chen, Jiaxi Li, Arif~Hassan Zidan, Afrar Jahin, Minheng Chen, Sichen Xia, Jason Holmes, Yan Zhuang, Jiaqi Wang, Bochen Xu, Weiran Xia, Jichao Yu, Kaibo Tang, Yaxuan Yang, Bolun Sun, Tao Yang, Guoyu Lu, Xianqiao Wang, Lilong Chai, He~Li, Jin Lu, Lichao Sun, Xin Zhang, Bao Ge, Xintao Hu, Lian Zhang, Hua Zhou, Lu~Zhang, Shu Zhang, Ninghao Liu, Bei Jiang, Linglong Kong, Zhen Xiang, Yudan Ren, Jun Liu, Xi~Jiang, Yu~Bao, Wei Zhang, Xiang Li, Gang Li, Wei Liu, Dinggang Shen, Andrea Sikora, Xiaoming Zhai, Dajiang Zhu, and Tianming Liu.
\newblock Evaluation of openai o1: Opportunities and challenges of agi, 2024.

\bibitem{qin2023toolllmfacilitatinglargelanguage}
Yujia Qin, Shihao Liang, Yining Ye, Kunlun Zhu, Lan Yan, Yaxi Lu, Yankai Lin, Xin Cong, Xiangru Tang, Bill Qian, Sihan Zhao, Lauren Hong, Runchu Tian, Ruobing Xie, Jie Zhou, Mark Gerstein, Dahai Li, Zhiyuan Liu, and Maosong Sun.
\newblock Toolllm: Facilitating large language models to master 16000+ real-world apis, 2023.

\bibitem{xi2023risepotentiallargelanguage}
Zhiheng Xi, Wenxiang Chen, Xin Guo, Wei He, Yiwen Ding, Boyang Hong, Ming Zhang, Junzhe Wang, Senjie Jin, Enyu Zhou, Rui Zheng, Xiaoran Fan, Xiao Wang, Limao Xiong, Yuhao Zhou, Weiran Wang, Changhao Jiang, Yicheng Zou, Xiangyang Liu, Zhangyue Yin, Shihan Dou, Rongxiang Weng, Wensen Cheng, Qi~Zhang, Wenjuan Qin, Yongyan Zheng, Xipeng Qiu, Xuanjing Huang, and Tao Gui.
\newblock The rise and potential of large language model based agents: A survey, 2023.

\bibitem{masterman2024landscapeemergingaiagent}
Tula Masterman, Sandi Besen, Mason Sawtell, and Alex Chao.
\newblock The landscape of emerging ai agent architectures for reasoning, planning, and tool calling: A survey, 2024.

\bibitem{yao2023reactsynergizingreasoningacting}
Shunyu Yao, Jeffrey Zhao, Dian Yu, Nan Du, Izhak Shafran, Karthik Narasimhan, and Yuan Cao.
\newblock React: Synergizing reasoning and acting in language models, 2023.

\bibitem{liu2024llmconversationalagentmemory}
Na~Liu, Liangyu Chen, Xiaoyu Tian, Wei Zou, Kaijiang Chen, and Ming Cui.
\newblock From llm to conversational agent: A memory enhanced architecture with fine-tuning of large language models, 2024.

\bibitem{shinn2023reflexionlanguageagentsverbal}
Noah Shinn, Federico Cassano, Edward Berman, Ashwin Gopinath, Karthik Narasimhan, and Shunyu Yao.
\newblock Reflexion: Language agents with verbal reinforcement learning, 2023.

\bibitem{guo2024embodiedllmagentslearn}
Xudong Guo, Kaixuan Huang, Jiale Liu, Wenhui Fan, Natalia Vélez, Qingyun Wu, Huazheng Wang, Thomas~L. Griffiths, and Mengdi Wang.
\newblock Embodied llm agents learn to cooperate in organized teams, 2024.

\bibitem{chen2023agentversefacilitatingmultiagentcollaboration}
Weize Chen, Yusheng Su, Jingwei Zuo, Cheng Yang, Chenfei Yuan, Chi-Min Chan, Heyang Yu, Yaxi Lu, Yi-Hsin Hung, Chen Qian, Yujia Qin, Xin Cong, Ruobing Xie, Zhiyuan Liu, Maosong Sun, and Jie Zhou.
\newblock Agentverse: Facilitating multi-agent collaboration and exploring emergent behaviors, 2023.

\bibitem{doi:10.1080/17538947.2024.2353122}
Siqin Wang, Tao Hu, Huang Xiao, Yun Li, Ce~Zhang, Huan Ning, Rui Zhu, Zhenlong Li, and Xinyue Ye.
\newblock Gpt, large language models (llms) and generative artificial intelligence (gai) models in geospatial science: a systematic review.
\newblock {\em International Journal of Digital Earth}, 17(1):2353122, 2024.

\bibitem{doi:10.1080/17538947.2023.2278895}
Zhenlong Li and Huan Ning.
\newblock Autonomous gis: the next-generation ai-powered gis.
\newblock {\em International Journal of Digital Earth}, 16(2):4668--4686, 2023.

\bibitem{doi:10.1080/15230406.2024.2404868}
Yifan Zhang, Zhengting He, Jingxuan Li, Jianfeng Lin, Qingfeng Guan, and Wenhao Yu.
\newblock Mapgpt: an autonomous framework for mapping by integrating large language model and cartographic tools.
\newblock {\em Cartography and Geographic Information Science}, 0(0):1--27, 2024.

\bibitem{doi:10.1080/17538947.2024.2398063}
Jianyuan Liang, Anqi Zhao, Shuyang Hou, Fengying Jin, and Huayi Wu.
\newblock A gpt-enhanced framework on knowledge extraction and reuse for geographic analysis models in google earth engine.
\newblock {\em International Journal of Digital Earth}, 17(1):2398063, 2024.

\bibitem{isprs-archives-XLVIII-4-W10-2024-113-2024}
H.~Kim and S.~Lee.
\newblock Poi gpt: Extracting poi information from social media text data.
\newblock {\em The International Archives of the Photogrammetry, Remote Sensing and Spatial Information Sciences}, XLVIII-4/W10-2024:113--118, 2024.

\bibitem{Xu2024GenAIpoweredMP}
Haowen Xu, Jinghui Yuan, Anye Zhou, Guanhao Xu, Wan Li, Xuegang Ban, and Xinyue Ye.
\newblock Genai-powered multi-agent paradigm for smart urban mobility: Opportunities and challenges for integrating large language models (llms) and retrieval-augmented generation (rag) with intelligent transportation systems.
\newblock {\em ArXiv}, abs/2409.00494, 2024.

\bibitem{Roberts2023GPT4GEOHA}
Jonathan Roberts, Timo Luddecke, Sowmen Das, K.~Han, and Samuel Albanie.
\newblock Gpt4geo: How a language model sees the world's geography.
\newblock {\em ArXiv}, abs/2306.00020, 2023.

\bibitem{komeili-etal-2022-internet}
Mojtaba Komeili, Kurt Shuster, and Jason Weston.
\newblock {I}nternet-augmented dialogue generation.
\newblock In Smaranda Muresan, Preslav Nakov, and Aline Villavicencio, editors, {\em Proceedings of the 60th Annual Meeting of the Association for Computational Linguistics (Volume 1: Long Papers)}, pages 8460--8478, Dublin, Ireland, May 2022. Association for Computational Linguistics.

\bibitem{heyueya2023solvingmathwordproblems}
Joy He-Yueya, Gabriel Poesia, Rose~E. Wang, and Noah~D. Goodman.
\newblock Solving math word problems by combining language models with symbolic solvers, 2023.

\bibitem{NEURIPS2023_d842425e}
Timo Schick, Jane Dwivedi-Yu, Roberto Dessi, Roberta Raileanu, Maria Lomeli, Eric Hambro, Luke Zettlemoyer, Nicola Cancedda, and Thomas Scialom.
\newblock Toolformer: Language models can teach themselves to use tools.
\newblock In A.~Oh, T.~Naumann, A.~Globerson, K.~Saenko, M.~Hardt, and S.~Levine, editors, {\em Advances in Neural Information Processing Systems}, volume~36, pages 68539--68551. Curran Associates, Inc., 2023.

\bibitem{sun2024verifiabletextgenerationevolving}
Hao Sun, Hengyi Cai, Bo~Wang, Yingyan Hou, Xiaochi Wei, Shuaiqiang Wang, Yan Zhang, and Dawei Yin.
\newblock Towards verifiable text generation with evolving memory and self-reflection, 2024.

\bibitem{bonatti2024windowsagentarenaevaluating}
Rogerio Bonatti, Dan Zhao, Francesco Bonacci, Dillon Dupont, Sara Abdali, Yinheng Li, Yadong Lu, Justin Wagle, Kazuhito Koishida, Arthur Bucker, Lawrence Jang, and Zack Hui.
\newblock Windows agent arena: Evaluating multi-modal os agents at scale, 2024.

\bibitem{githubGitHubSignificantGravitasAutoGPT}
{G}it{H}ub - {S}ignificant-{G}ravitas/{A}uto{G}{P}{T}: {A}uto{G}{P}{T} is the vision of accessible {A}{I} for everyone, to use and to build on. {O}ur mission is to provide the tools, so that you can focus on what matters. --- github.com.
\newblock \url{https://github.com/Significant-Gravitas/AutoGPT}.
\newblock [Accessed 09-10-2024].

\bibitem{2017-ak}
{Guoan Tang}.
\newblock {\em 100 Basic Experiments in GIS (Geographic Information System)}.
\newblock Science Press, 2017.

\bibitem{patil2023gorillalargelanguagemodel}
Shishir~G. Patil, Tianjun Zhang, Xin Wang, and Joseph~E. Gonzalez.
\newblock Gorilla: Large language model connected with massive apis, 2023.

\end{thebibliography}


\end{document}